\documentclass[smallcondensed]{svjour3} 
\usepackage{url}
\usepackage{graphicx}
\usepackage[cmex10]{amsmath}
\usepackage{amssymb}
\usepackage{array}
\usepackage{algorithm,algpseudocode}
\usepackage{mathtools}
\usepackage{longtable}
\usepackage{lscape,caption}
\usepackage{booktabs}
\usepackage{color}
\usepackage{mathtools}
\usepackage{subfigure} 
\usepackage[numbers]{natbib}

\RequirePackage[colorlinks,citecolor=blue,urlcolor=black,linkcolor=black]{hyperref}
\usepackage[misc,geometry]{ifsym}
\usepackage[labelfont=bf]{caption}

\usepackage{etoolbox}
 
\newcommand*{\affmark}[1][*]{\textsuperscript{#1}}
\newcommand\tab[1][0.5cm]{\hspace*{#1}}

\begin{document}

\title{AEkNN: An AutoEncoder kNN-based classifier with built-in dimensionality reduction}

\titlerunning{AEkNN: An AutoEncoder kNN-based classifier ...}

\author{Francisco J. Pulgar\affmark[1] \href{http://orcid.org/0000-0003-3558-3990}{\includegraphics[scale=0.05]{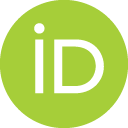}}  \and Francisco Charte\affmark[1]\href{https://orcid.org/0000-0002-3083-8942}{\includegraphics[scale=0.05]{Figures/orcid.png}}\and Antonio J. Rivera\affmark[1]\href{http://orcid.org/0000-0002-1062-3127}{\includegraphics[scale=0.05]{Figures/orcid.png}} \and Mar\'ia J. del Jesus\affmark[1]\href{http://orcid.org/0000-0002-7891-3059}{\includegraphics[scale=0.05]{Figures/orcid.png}}}

\institute{\Letter\tab  F. J. Pulgar  \at
	\phantom{\Letter}\tab    Tel.: +34 953 211 956\\
	\phantom{\Letter}\tab \href{mailto:fpulgar@ujaen.es}{fpulgar@ujaen.es}\\
	\phantom{\Letter}\tab  F. Charte  \at
	\phantom{\Letter}\tab \href{mailto:fpulgar@ujaen.es}{fcharte@ujaen.es}\\
	\phantom{\Letter}\tab  A. J. Riverea  \at
	\phantom{\Letter}\tab \href{mailto:fpulgar@ujaen.es}{arivera@ujaen.es}\\
	\phantom{\Letter}\tab  M. J. del Jesus  \at
	\phantom{\Letter}\tab \href{mailto:fpulgar@ujaen.es}{mjjesus@ujaen.es}\\\\
	1 \tab Department of Computer Science, University of Ja\'en, 23071 Ja\'en, Spain \\}

\authorrunning{F. J. Pulgar et al. }

\date{}

\maketitle
\vspace*{-1cm}
\begin{abstract}
High dimensionality, i.e. data having a large number of variables, tends to be a challenge for most machine learning tasks, including classification. A classifier usually builds a model representing how a set of inputs explain the outputs. The larger is the set of inputs and/or outputs, the more complex would be that model. There is a family of classification algorithms, known as lazy learning methods, which does not build a model. One of the best known members of this family is the kNN algorithm. Its strategy relies on searching a set of nearest neighbors, using the input variables as position vectors and computing distances among them. These distances loss significance in high-dimensional spaces. Therefore kNN, as many other classifiers, tends to worse its performance as the number of input variables grows.

In this work AEkNN, a new kNN-based algorithm with built-in dimensionality reduction, is presented. Aiming to obtain a new representation of the data, having a lower dimensionality but with more informational features, AEkNN internally uses autoencoders. From this new feature vectors the computed distances should be more significant, thus providing a way to choose better neighbors. A experimental evaluation of the new proposal is conducted, analyzing several configurations and comparing them against the classical kNN algorithm. The obtained conclusions demonstrate that AEkNN offers better results in predictive and runtime performance.


\end{abstract}

\section{Introduction}
Classification is a well-known task within the area of machine learning \cite{MachineLearning}. The main objective of a classifier is to find a way to predict the label to be assigned to new data patterns. To do so, usually a model is  created from previously labeled data. In traditional classification, each example has a single label. Different algorithms have been proposed in order to address this work. One of the classic methodologies is instance-based learning (IBL) \cite{aha1991}. Essentially, this methodology is based on local information provided by the training instances, instead of constructing a global model from the whole data. The algorithms belonging to this family are relatively simple, however they have demonstrated to obtain very good results in facing the classification problem. A traditional example of such an algorithm is k-nearest neighbors (kNN) \cite{cover1967}. 

The different IBL approaches, including the kNN algorithm \cite{Beyer1999}, have difficulties when faced with high-dimensional datasets. These datasets are made of samples having a large number of features. In particular, the kNN algorithm presents problems when calculating distances in high-dimensional spaces. The main reason is that distances are less significant as the number of dimensions increases, tending to equate \cite{Beyer1999}. This effect is one of the causes of the curse of dimensionality, which occurs when working with high-dimensional data \cite{Richard1957,Richard1961}. Another consequence that emerges in this context is the Hughes phenomenon. This fact implies that the predictive performance of a classifier decreases as the number of features of the dataset grows, keeping the number of examples constant \cite{hughes}. In other words, more instances would be needed to maintain the same level of performance.

Several approaches have been proposed for facing the dimensionality reduction task. Recently, a few proposals based on deep learning (DL) \cite{Deng2014,Bengio2013} have obtained good results while tackling this problem. The rise of these techniques is produced by the good performance that DL models have had in many research areas, such as computer vision, automatic speech processing, or audio and music recognition. In particular, autoencoders (AEs) are DL networks offering good results due to their architecture and operation \cite{CHARTE201878,hinton2006,Wang2014,Rifai2011,Zhou2016,Bhatia2018}.

High dimensionality is usually mitigated by transforming the original input space into a lower-dimensional one. In this paper an instance-based algorithm, that internally generates a reduced set of features, is proposed. Its objective is to obtain a better IBL method, able to deal with high-dimensional data. Specifically, the present work introduces AEkNN, a kNN-based algorithm with built-in dimensionality reduction. AEkNN projects the training patterns into a lower-dimensional space, relying in an AE for doing so. The goal is to produce new features of higher quality from the input data. This approach is experimentally evaluated, and a comparison between AEkNN and the kNN algorithm is performed considering predictive performance and execution time. The results obtained demonstrate that AEkNN offers better results in both metrics. In addition, AEkNN is compared with other traditional dimensionality reduction algorithms. This comparison offers an idea of the behavior of the AEkNN algorithm when facing the task of dimensionality reduction.

An important aspect of the AEkNN algorithm that must be highlighted is that it performs a transformation of the features of the input data, against other traditional algorithms that perform a simple selection of the most significant features. AEkNN performs this transformation taking into account all the characteristics of the input data, although not all will have the same weight in the new generated space.

In short, AEkNN combines two reference methods, kNN and AE, in order to take advantage of kNN in classification and reduce the effects of high dimensionality by means of AE. In this way, the proposed method presents a baseline for future works. Summarizing, the main contributions of this work are 1) the design of a new classification algorithm, AEkNN, which combines an efficient dimensionality reduction mechanism with a popular classification method, 2) an analysis of the AEkNN operating parameters that allows selecting the best algorithm configuration, 3) an experimental demonstration of the improvement that AEkNN achieves with respect to the kNN algorithm, and 4) an experimental comparison between the use of an autoencoder for dimensionality reduction with respect to other classical methods such as PCA and LDA.

This paper is organized as follows. In Section \ref{Preliminaries} a few fundamental concepts, such as machine learning, classification, and the kNN algorithm, are briefly introduced, and some details about DL techniques and AEs are provided. Section \ref{RelatedWork} describes relevant related works, focused on tackling the problem of dimensionality reduction in kNN. In Section \ref{Proposal}, the proposed AEkNN  algorithm is introduced. Section \ref{Experimentation} defines the experimental framework, as well as the different results obtained from the experimentation. Finally, Section \ref{Concluding} provides the overall conclusions.

\section{Preliminaries}\label{Preliminaries}
AEkNN, the algorithm proposed in this work, is a kNN-based classification method designed to deal with high-dimensional data. This section outlines the essential concepts AEkNN is founded on, such as classification, nearest neighbors classification, DL techniques and AEs. The most basic concepts are introduced in subsection \ref{Classification}. The kNN algorithm is discussed in subsection \ref{KNN}, while DL and AEs are briefly described in subsections \ref{DeepLearning} and \ref{Autoencoder}.

\subsection{Machine Learning and Classification}\label{Classification}
In general terms, machine learning is a subfield of artificial intelligence whose aim is to develop algorithms and techniques able to generalize behaviors from information supplied as examples \cite{kohavi1998,Goldberg1988}. The different tasks that can be performed on machine learning can be classified following different criteria. One of these categorizations arises according to the training patterns used to train the machine learning system \cite{Alpa2014}. In this sense, it can be distinguished between supervised learning, when patterns are labeled, such as classification and regression, and unsupervised learning, when patterns are not labeled, such as clustering, among others.

Classification is one of the tasks performed in the data mining phase. It is a predictive task that usually develops through supervised learning methods \cite{kotsiantis}. Its purpose is to predict, based on previously labeled data, the class to assign to future unlabeled patterns. In traditional classification, datasets are structured as a set of input attributes, known as features, and one output attribute, the class or label. Depending on the number of values that this output class can take, the classification problem can be seen as:

\begin{itemize}
	\item Binary classification, in which each pattern can only belong to one of two classes. 
	
	\item Multi-class classification, in which each pattern can only belong to one of a limited set of classes. Consequently, a binary classification can be seen as a problem of multi-class classification with only two classes.
\end{itemize}

Several issues can emerge while designing a classifier, being some of them related to high dimensionality. According to the Hughes phenomenon \cite{hughes}, the predictive performance of a classifier decreases as the number of features increases, provided that  the number of training instances is constant. Another phenomenon that particularly affects IBL algorithms is the curse of dimensionality. IBL algorithms are based on the similarity of individuals, calculating distances between them \cite{aha1991}. These distances tend to lose significance as dimensionality grows.

\subsection{The kNN Algorithm}\label{KNN}
\begin{figure*}[h]
	\centering
	\includegraphics[width=0.5\textwidth]{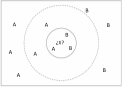}
	\caption{k-Nearest neighbors algorithm in a bi-dimensional space.} \label{knn}
\end{figure*}

kNN is a non-parametric algorithm developed to deal with classification and regression tasks\cite{altman,cover1967}. In classification, kNN predicts the class for new instances using the information provided by the \textit{k} nearest neighbors, so that the assigned class will be the most common among them. Fig. \ref{knn} shows a very simple example on how kNN works with different \textit{k} values. As can be seen, the prediction obtained with \textit{k} = 3 would be \textit{B}, with \textit{k} = 5 would be \textit{A} and with \textit{k} = 11 would be \textit{A}.

An important feature of this algorithm is that it does not build a model for accomplishing the prediction task. Usually, no work is done until an unlabeled data pattern arrives, thus the denomination of \textit{lazy} approach \cite{Atkeson1997}. Once the instance to be classified is given, the information provided by its \textit{k} nearest neighbors \cite{dasarathy1991} is used as explained above. 

One of kNN's main issues is its behavior with datasets having a high-dimensional input space, due to the loss of significance of traditional distances as the dimensionality of the data increases \cite{Beyer1999}. In such a high-dimensional space distances between individuals tend to be the same. As a consequence similarity/distance-based algorithms, such as kNN, usually do not offer adequate results.

In this article, kNN has been selected to perform classification tasks. kNN is very popular since it has a good performance, uses few resources and it is relatively simple \cite{maillo2017knn,xia2017demst,maillo2017exact,deng2016efficient}. The objective of this proposal is to present an algorithm that combines the advantages of kNN to classify with DL models to reduce dimensionality.

\subsection{Deep Learning}\label{DeepLearning}
DL \cite{willey2016,Deng2014} arises with the objective of extracting higher-level representations of the analyzed data. In other words, the main goal of DL-based techniques is learning complex representations of data. The main lines of research in this area are intended to define different data representations and create models to learn them \cite{bengio2013_1}.

As the name suggests, models based on DL are developed as multi-layered (deep) architectures, which are used to map the relationships between features in the data (inputs) and the expected result (outputs) \cite{goodfellow2016,bengio2009}. Most DL algorithms learn multiple levels of representations, producing an hierarchy of concepts. These levels correspond to different degrees of abstraction. The following are some of the main advantages of DL: 

\begin{itemize}
	\item These models can handle a large number of variables and generate new features as part of the algorithm itself, not as an external step \cite{Deng2014,image2,Tang2017}. 
	
	\item Provides performance improvements in terms of time needed to accomplish feature engineering, one of the most time-consuming tasks \cite{goodfellow2016}.
	
	\item Achieves much better results than other methods based on traditional techniques \cite{image1,image3,speech2,Ye2018} while facing problems in certain fields, such as image, speech recognition or malware detection.
	
	\item DL-based models have a high capacity of adaptation for facing new problems \cite{willey2016,bengio2009}.
\end{itemize}

Recently, several new methods \cite{Bengio2013,LeCun2015,Deng2014} founded on the good results produced by DL have been published. Some of them are focused on certain areas, such as image processing and voice recognition \cite{LeCun2015}. Other DL-based proposals have been satisfactorily applied in disparate areas, gaining advantage over prior techniques \cite{Deng2014}. Due to the great impact of DL-based techniques, as well as the impressive results they usually offer, new challenges are also emerging in new research lines \cite{Bengio2013}.

There are two main reasons behind the rise of DL techniques, the large amount of data currently available and the increase in processing power. In this context, different DL architectures have been developed: AEs (section \ref{Autoencoder}), convolutional neural networks \cite{LeCun1995}, long short-term memory \cite{hoch1997}, recurrent neural networks \cite{sak2014}, gated recurrent unit \cite{chung2015}, deep Boltzmann machines \cite{hinton2012}, deep stacking networks \cite{deng2011}, deep coding networks \cite{lin2010deep}, deep recurrent encoder \cite{sordoni2015hierarchical}, deep belief networks \cite{Keyvanrad2017,Lee2009}, among others \cite{Deng2014,bengio2009}.

DL models have been widely used to perform classification tasks obtaining good results \cite{image1,image3,esteva2017dermatologist,deng2017hierarchical}. However, the objective of this proposal is not to perform the classification directly with these models, but use them to the dimensionality reduction task.

The goal of the present work is to obtain higher-level representations of the data but with a reduced dimensionality. One of the dimensionality reduction DL-based techniques that has achieved good results are AEs \cite{liou2014}. An AE is an artificial neural network whose purpose is to reproduce the input into the output, in order to generate a compressed representation of the original information \cite{bengio2009}. Section \ref{Autoencoder} thoroughly describes this technique.

\subsection{Autoencoders}\label{Autoencoder}

An AE is an artificial neural network able to learn new information encodings through unsupervised learning \cite{CHARTE201878}. AEs are trained to learn an internal representation that allows reproducing the input into the output through a series of hidden layers. The goal is to produce a more compressed representation of the original information in the hidden layers, so that it can be used in other tasks. AEs are typically used for dimensionality reduction tasks by their characteristics and performance \cite{hinton2006,Wang2014,Rifai2011,Zhou2016,Bhatia2018,snoek2012nonparametric,zhang2014supervised}. Therefore, the importance of such networks in this paper.

The most basic structure of an AE is very similar to that of a multilayer perceptron. An AE is a feedforwark neural network without cycles, so the information always goes in the same direction. An AE is typically formed by a series of layers: an input layer, a series of hidden layers and an output layer, being the units in each layer connected to those in the next one. The main characteristic of AEs is that the output has the same number of nodes than the input, since the goal is to reproduce the latter into the former throughout the learning process \cite{bengio2009}.

\begin{figure*}[h!]
	\centering
	\includegraphics[width=0.75\textwidth]{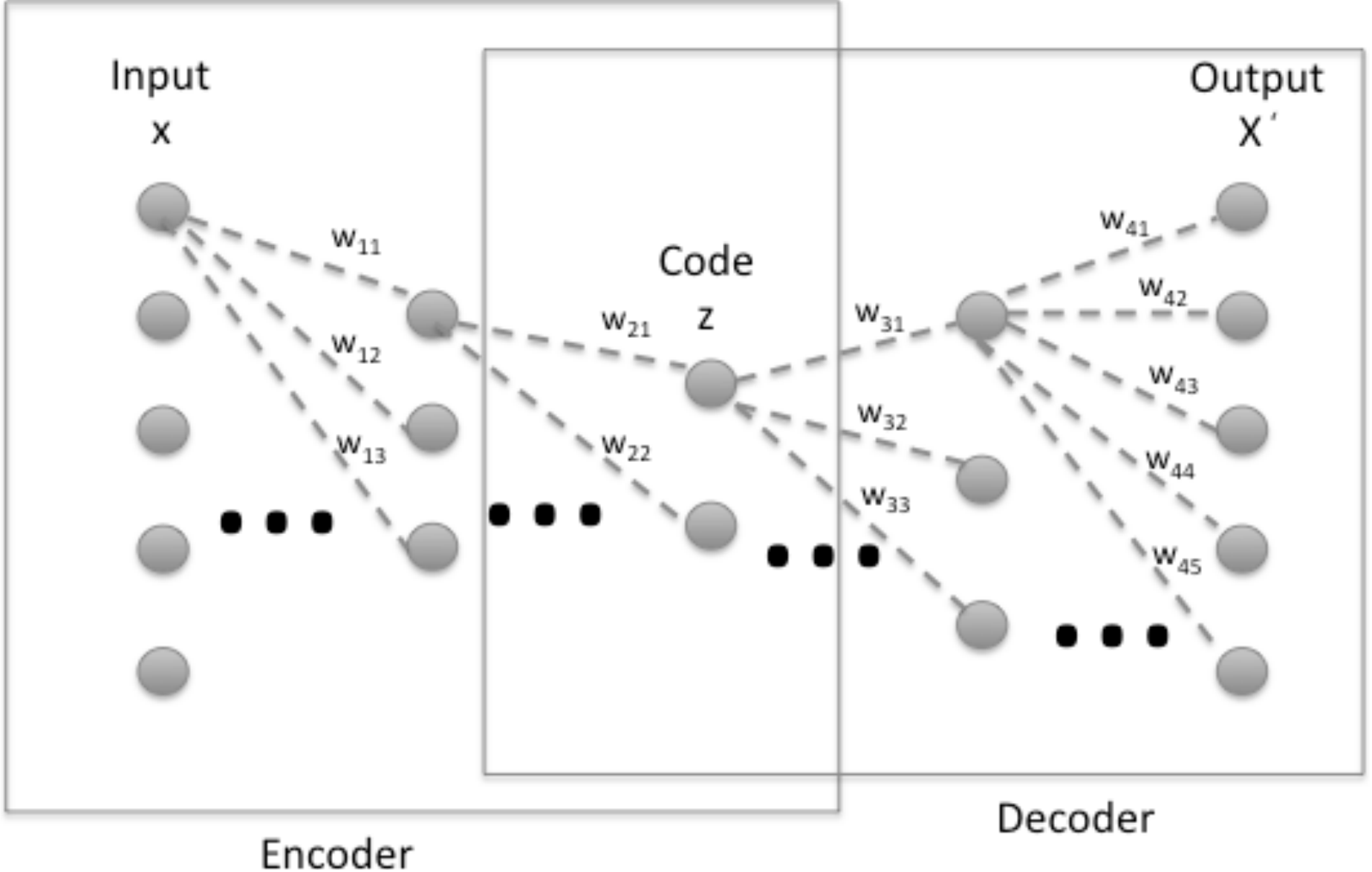}
	\caption{Architecture of an autoencoder with three hidden layers.} \label{Autoencoder_structure}
\end{figure*}

Two parts can be distinguished in an AE, the encoder and the decoder. The first one is made up of the input layer and the first half of hidden layers. The second is composed of the second half of hidden layers and the output layer. This is the architecture shown in Fig. \ref{Autoencoder_structure}. As can be seen, the structure of an AE always is symmetrical.

The encoder and decoder parts in an AE can be defined as functions $\omega$ (Eq. (\ref{A1})) and $\beta$ (Eq. (\ref{A2})), so that:

\begin{equation}
\omega : X \rightarrow F \label{A1}
\end{equation}

\begin{equation}
\beta : F \rightarrow X \label{A2}
\end{equation}
%\begin{equation}
%\argmin\limits_{\mathbf{\omega,\beta}}\left \| X - (\omega\;o\;\beta) X \right \|^{2}
%\end{equation}

Where $x \in \mathbb{R}^{d} = X $ is the input to the AE, and $z \in \mathbb{R}^{p} = F $ is the mapping contained in the hidden layers of the AE. When there is only one hidden layer (the most basic case) the AE maps the input \textit{X} onto \textit{Z}. For doing this, a weight vector \textit{W} and a bias parameter \textit{b} are used:

\begin{equation}
z\;=\; \gamma_{1}\;(W\;x\;+\;b) \label{A3}
\end{equation}

Eq. (\ref{A3}) corresponds to the compression function, wherein a encoded input representation is obtained. Here, $\gamma_{1}$ is an activation function such as a rectified linear unit or a sigmoid function. 

The next step is to decode the new representation \textit{z} to obtain \textit{x'}, in the same way as \textit{x} was projected into \textit{z}, by means of a weight vector \textit{W'} and a bias parameter \textit{b'}. Eq. (\ref{A4}) corresponds to the decoder part, where the AE reconstructs the input from the information contained in the hidden layer.

\begin{equation}
\textit{x'}\;=\; \gamma_{2}\;(W'\;z\;+\;b') \label{A4}
\end{equation}

During the training process, the AE tries to reduce the reconstruction error. This operation consists of back-propagating the obtained error through the network, and then modifying the weights to minimize such error. Algorithm \ref{AEAlg} shows the pseudo-code of this process.
%(Equation \ref{A5})
%\begin{equation}
%L(x,\;x')\;=\; \left \| x\;-\;x' \right \|^{2}\;=\;\left \| x\;-\; %\gamma_{2}\;(W'\;\gamma_{1}\;(W\;x\;+\;b)\;+\;b')\right \|^{2} \label{A5}
%\end{equation}

\algnewcommand{\LineComment}[1]{\State \(\triangleright\) #1}
\begin{algorithm}[h]
	\small
	%\footnotesize 
	\caption{AE training algorithm's pseudo-code.}
	\label{AEAlg}
	\begin{algorithmic}[1]
		\Statex \textbf{Inputs}:
		\Statex	\hspace{3em}$TrainData$ \Comment Train Data
	\end{algorithmic}
	\begin{algorithmic}[1]
		\LineComment{For each training instance:}
		\For{\textbf{each} $instace$ \textbf{in} $TrainData$} 
		\LineComment{Do a feed-forward through AE to obtain output:}
		\State \textit{newCod} $\gets$ feedForwardAE($aeModel,instance$)
		\LineComment{Calculate error:}
		\State \textit{error} $\gets$ calculateError($newCod, instance$)
		\LineComment{Backpropagate the error through AE and perform weight update:}
		\State \textit{aeModel} $\gets$ backpropagateError($eror$)
		\EndFor
	\end{algorithmic}
\end{algorithm}

Learning a representation that allows reproducing the input into the output could seem useless at first sight, but in this case the output is not of interest. Instead, the concern is in the new representation of the inputs learned in the hidden layers. Such new codification is really interesting, because it can have very useful features \cite{goodfellow2016}. The hidden layers learn a higher-level representation of the original data, which can be extracted and used in independent processes.

Depending on the number of units the hidden layers have, two types of AEs can be distinguished:

\begin{itemize}
	\item Those whose hidden layer have less units than the input and output layers. This type of AE is called undercomplete. Its main objective is to force the network to learn a compressed representation of the input, extracting new, higher-level features.
	
	\item Those whose hidden layer have more units than the input and output layers. This type of AE is called overcomplete. The main problem in this case is that the network can learn to copy the input to the output without learning anything useful, so when it is necessary to obtain an enlarged representation of the input it is necessary to use other tools to prevent this problem.
\end{itemize}

In conclusion, AEs are a very suitable tool for generating a new lower-dimensional input space made of higher-level features. AEs have obtained good results in the accomplishment of this task. This is the main reason to choose this technique to design the AEkNN algorithm described later. However, it is important to note that there are other methods of dimensionality reduction that produce good results, such as denoising autoencoders \cite{vincent2008extracting}, restricted Boltzmann machines \cite{salakhutdinov2007restricted} or sparse coding \cite{yang2011robust}. The objective of this proposal is to present an algorithm that hybridizes kNN with AEs. This establishes a baseline that allows supporting studies with more complex methods.

\section{Dimensionality reduction approaches}\label{RelatedWork}

In this section, an exploration of previous works related to the proposal made in this paper is carried out. The subsection \ref{Clasical Dimensionality Reduction} introduces classical proposals to tackle the dimensionality reduction problem. Some approaches for facing the dimensionality reduction task for kNN are portrayed in subsection \ref{KNN Dimensionality Reduction}.

In automatic learning, dimensionality reduction is the process aimed to decrease the number of considered variables, by obtaining a subset of main features. Usually two different dimensionality reduction methods are considered:

\begin{itemize}
	\item Feature selection \cite{liu2007}, where the subset of the initial features that provides more useful information is chosen. The final features have no transformation in the process.
	
	\item Feature extraction \cite{liu1998}, where the process constructs from the initial features a set of new ones  providing more useful and non-redundant information, facilitating the next steps in machine learning and in some cases improving understanding by humans.
\end{itemize}

\subsection{Classical proposals for dimensionality reduction}\label{Clasical Dimensionality Reduction}
Most dimensionality reduction techniques can be grouped into two categories, linear and non-linear approaches \cite{Laurens2009}. Below some representative proposals found in the literature, those that can be considered as traditional methods, are depicted.

Commonly, classic proposals for dimensionality reduction were developed using linear techniques. The following are some of them:

\begin{itemize}
	\item Principal Components Analysis (PCA) \cite{Pearson1901,Hotelling1933} is a well-known solution for dimensionality reduction. Its objective is to obtain the lower-dimensional space where the data are embedded. In particular, the process starts from a series of correlated variables and converts them into a set of uncorrelated linear variables. The variables obtained are known as principal components, and their number is less than or equal to the original variables. Often, the internal structure extracted in this process reflects the variance of the data used.
	
	\item Factors analysis \cite{Spearman1904} is  based on the idea that the data can be grouped according to their correlations, i.e. variables with a high correlation will be within the same group and variables with low correlation will be in different groups. Thus, each group represents a factor in the observed correlations. The potential factors plus the terms error are linearly combined to model the input variables. The objective of factor analysis is to look for independent dimensions.
	
	\item Classical scaling \cite{Torgerson1952} consists in grouping the data according to their similarity level. An algorithm of this type aims to place each data in an N-dimensional space where the distances are maintained as much as possible.
	
\end{itemize}

Despite their popularity, classical linear solutions for dimensionality reduction present the problem that they can not correctly handle complex non-linear data \cite{Laurens2009}. For this reason, nonlinear proposals for dimensionality reduction arose. A compilation of these is presented in \cite{burges2005,saul2006,Lee2007,Venna2007}. Some of these techniques are: Isomap \cite{tenenbaum2000}, Maximum Variance Unfolding \cite{weinberger2006}, diffusion maps \cite{coifman2005}, manifold charting \cite{Laurens2009}, among others. These techniques allow to work correctly with complex non-linear data. This is an advantage when working with real data, which are usually of this type.

\subsection{Proposals for dimensionality reduction in kNN}\label{KNN Dimensionality Reduction}

There are different proposals trying to face the problems of kNN when working with high-dimensional data. In this section, some of them are collected:

\begin{itemize}
	\item A method for computing distances for kNN is presented in \cite{yu2001}. The proposed algorithm performs a partition of the total data set. Subsequently, a reference value for each partition made is assigned. Grouping the data in different partitions allows to obtain a space of smaller dimensionality where the distances between the reference points are more significant. The method depends on the division of the data performed and the selection of the reference. This is a negative aspect, since a poor choice of parameters can greatly influence the final results.
	
	\item The authors of \cite{kou2011} analyze the curse of dimensionality phenomenon, which states that in high-dimensional spaces the distances between the data points become almost equal. The objective is to investigate when the different methods proposed reach their limits. To do this, they perform an experiment in which different methods are compared with each other. In particular, it is exposed that the kNN algorithm begins to worsen the results when the space exceeds eight dimensions. A proposed solution is to adapt the calculation of distances to high-dimensional spaces. However, this approach does not consider a transformation of the initial data to a lower-dimensional space.
	
	\item The proposal in \cite{wang2012} is a kNN-based method called kMkNN, whose objective is to improve the search of the nearest neighbors in a high-dimensional space. This problem is approached from another point of view, being the goal to accelerate the computation of distances between elements without modifying the input space. To do this, kMkNN relies on k-means clustering and the triangular inequality. The study shows a comparison with the original kNN algorithm where it is demonstrated that kMkNN works better considering the execution time, although it is not as effective when predictive performance is taken into account. 
	
	\item A new aspect related to the curse of dimensionality phenomenon, occurring while working with the kNN algorithm, is explored in \cite{Rado2010}. It refers to the number of times that a particular element appears as the closest neighbor of the rest of elements. The main objective of the study is to examine the origins of this phenomenon as well as how it can affect dimensionality reduction methods. The authors analyze this phenomenon and some of its effects on kNN classification, providing a foundation which allows making  different proposals aimed to mitigate these effects.
	
	\item In \cite{hinne2000}, the problem of finding the nearest neighbors in a high-dimensional space is analyzed. This is a difficult problem both from the point of view of performance and the quality of results. In this study, a new search approach is proposed, where the most relevant dimensions are selected according to certain quality criteria. Thus, the different dimensions are not treated in the same way. This can be seen as an extraction of characteristics according to a particular criterion. Finally, an algorithm based on the previous approach, that faces the problem of the nearest neighbor, is proposed. However, this method makes a selection of the initial features that meet a certain criterion, so it does not take into account all the input features. Therefore, it could discard important information in the process.
	
	\item A method called DNet-kNN is presented in \cite{Min2009}. It consists in a non-linear feature mapping based on Deep Neural Network, aiming to improve classification by kNN. DNet-kNN relies on Restricted Boltzmann Machines (RBM) to perform the mapping of characteristics. RBMs are another type of DL network. The work offers a solution to the problem of high-dimensional data when using kNN by combining this traditional classifier with DL-based techniques. The conducted experimentation proves that DNet-kNN improves the results of the classical kNN. DNet-kNN requires a pre-training of the RBM, which is an additional phase. In addition, experimentation is performed on datasets of digits and letters, not on actual images.
\end{itemize}

The aforementioned proposals represent a list of approaches that are closely related to the problem dealt with in this article. However, there are many other proposals that face the problem of dimensionality reduction from other perspectives \cite{roweis2000nonlinear,belkin2003laplacian,casillas2001genetic,harsanyi1994hyperspectral,yu2001direct}.

In conclusion, different proposals have arisen to analyze and try to address the problems of IBL algorithms when they have to deal with high-dimensional data. These methods are affected by the curse of dimensionality, which raises the need to bring new approaches. Among the previous proposals, there is no one that presents an hybrid method based on IBL that incorporates the reduction of dimensionality intrinsically. In addition, none of the proposals obtains improvements both in predictive performance as well as in execution time. The present work is aimed to fulfill these aspects.

\section{AEkNN: An AutoEncoder kNN-based classifier with built-in dimensionality reduction}\label{Proposal}
Once the main foundations behind our proposed algorithm have been described, this section proceeds to present AEkNN. It is an instance-based algorithm with a built-in dimensionality reduction method, performed with DL techniques, in particular by means of AEs.

\subsection{AEkNN Foundations}
As mentioned, high dimensionality is an obstacle for most classification algorithms. This problem is more noticeable while working with distance-based methods, since as the number of variables increases these distances are less significant \cite{hughes,kou2011}. In such situation an IBL method could loss effectiveness, since the distances between individuals are equated. As a consequence of this problem, different methods able to reduce its effects have been proposed. Some of them have been previously enumerated in Section \ref{KNN Dimensionality Reduction}.

AEkNN is a new approach in the same direction, introducing the use of AEs to address the problem of high dimensionality. The structure and performance of this type of neural networks makes it suitable for this task. As explained above, AEs are trained to reproduce the input into the output, through a series of hidden layers. Usually, the central layer in AEs has less units than the input and output layers. Thereby, this bottleneck forces the network to learn a more compact and higher-level representation of the information. The coding produced by this central layer can be extracted and used as the new set of features in the classification stage. In this sense, there are different studies demonstrating that better results are obtained with AEs than with traditional methods, such PCA or multidimensional scaling \cite{hinton2006, Laurens2009}. Also, there are studies analyzing the use of AEs from different perspectives, either focusing on the training of the network and its parameters \cite{Wang2014} or on the relationship between the data when building the model \cite{Rifai2011}.

AEkNN is an instance-based algorithm designed to address classification with a high number of variables. It starts working with an \textit{N}-dimensional space \textit{X} that is projected into an \textit{M}-dimensional space \textit{Z}, being \textit{M $<$ N}. This way \textit{M} new features, presumably of higher level than the initial ones, are obtained. Once the new representation of the input data is generated, it is possible to get more representative distances. To estimate the output, the algorithm uses the distances between each test example and the training ones but based on the \textit{M} higher-level features. Thus, the drawbacks of high-dimensional data in  distances computation can be significantly reduced. As can be seen, AEkNN is a non-lazy instance-based algorithm. It starts by generating the model in charge of producing the new features, unlike the lazy methods that do not have a learning stage or model. AEkNN allows to enhance the predictive performance, as well as obtaining improvements in execution time, when working with data having a large number of features.

\subsection{Method description}
\begin{figure*}[h!]
	\centering
	\includegraphics[width=0.75\textwidth]{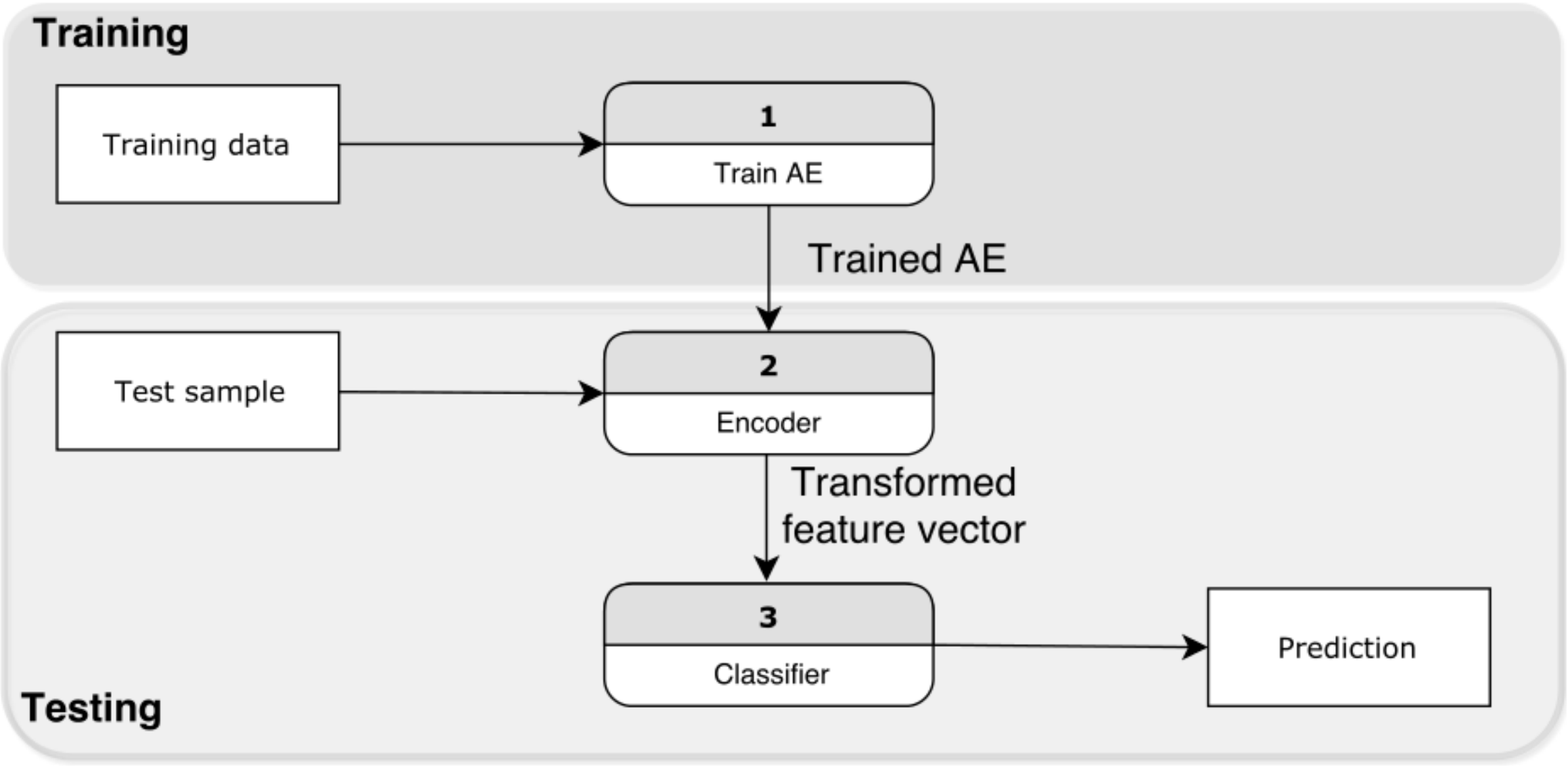}
	\caption{Method description.} \label{AEkNN}
\end{figure*}

AEkNN consists of two fundamental phases. Firstly, the learning stage is carried out using the training data to generate the AE model that allows to produce a new encoding of the data. Secondly, the classification step is performed. It uses the model generated in the first phase to obtain the new representation of the test data and, later, the class for each instance is estimated based on nearest neighbors. Algorithm \ref{AEkNNAlg} shows the pseudo-code of AEkNN, that is thoroughly discussed below, while Fig. \ref{AEkNN} shows the algorithm process in a general way.

\begin{algorithm}[h!]
	\small
	%\footnotesize 
	\caption{AEkNN algorithm's pseudo-code.}
	\label{AEkNNAlg}
	\begin{algorithmic}[1]
		\Statex \textbf{Inputs}:
		\Statex	\hspace{3em}$TrainData$ \Comment Train Data
		\Statex	\hspace{3em}$TestData$ \Comment Test Data
		\Statex	\hspace{3em}$PPL$ \Comment Percentage per layer	 
		\Statex	\hspace{3em}\textit{k} \Comment Number of nearest neighbors
	\end{algorithmic}
	
	\begin{algorithmic}[1]
		\LineComment{Training phase:}
		\State \textit{modelData} $\gets$ \textit{TrainData} 
		\State \textit{aeModel} $\gets$ \textit{$($$)$} 
		\For{\textbf{each} $layer$ \textbf{in} \textit{PPL}} 
		\State \textit{sizeLayer} $\gets$ getSizeLayer($modelData, layer$)
		\State \textit{aeLayer} $\gets$ getAELayer($modelData, sizeLayer$)
		\State \textit{modelData} $\gets$ applyAELayer($aeLayer,modelData$) 
		\State \textit{aeModel} $\gets$ addAEModel($aeModel,aeLayer$) 
		\EndFor
		\LineComment{Classification phase:}
		\State \textit{result} $\gets$ classification($TrainData, TestData, k, aeModel$)	
		\State \textbf{return} \textit{result}
		\State
		\Function{getAELayer}{\textit{modelData}, \textit{sizeLayer}}
			\State \textit{aeLayer} $\gets$ initializeAEModel($modelData, sizeLayer$)
			\For{\textbf{each} $instance$ \textbf{in} $modelData$} 
			\State \textit{outPut} $\gets$ feedForwardAEModel($aeLayer, instance$)
			\State \textit{error} $\gets$ calculateDeviation($instance, outPut$)
			\State \textit{aeLayer} $\gets$ updateWeightsAEModel($aeLayer, error$)
			\EndFor
			\State \textbf{return} \textit{aeLayer}
		\EndFunction
		\State
		\Function{classification}{\textit{TrainData}, \textit{TestDAta},\textit{k},\textit{aeModel}}
		\State \textit{error} $\gets$ 0
		\For{\textbf{each} $instace$ \textbf{in} $TestData$} 
		\State \textit{newCod} $\gets$ feedForwardAEModelVector($aeModel,instance$)
		\State \textit{outPut} $\gets$ distanceBasedClassification($newCod, k, TrainData$)
		\If{ $outPut$ \textbf{!=} $realOutPut(instance)$} 
		\State \textit{error} $\gets$ error + 1
		\EndIf
		\EndFor
		\State \textit{result} $\gets$ error / size($TestData$)
		\State \textbf{return} \textit{result}
		\EndFunction
	\end{algorithmic}
\end{algorithm}

The inputs to the algorithm are \textit{TrainData} and \textit{TestData}, the train and test data to be processed, \textit{k}, the number of neighbors, and \textit{PPL}, the percentage of elements per layer (PPL). This latter parameter sets the structure of the AE, i.e. the number of layers and elements per layer. It is a vector made of as many items as hidden layers are desired in the AE, indicating each one the percentage of units in that layer with respect to the number of input features. In section \ref{Experimentation}, different configurations are analyzed to find the one that offers the best results.

The algorithm is divided into two parts. The first part of the code (lines 2-9) corresponds to the training phase of AEkNN. The second part (lines 11-12) refers to the classification phase. During training AEkNN focuses on learning a new representation of the data. This is done through an AE, using the training data to learn the weights linking the AE's units. This is a process that has to be repeated for each layer in the AE, stated by the number of elements in the \textit{PPL} parameter. This loop performs several tasks:

\begin{itemize}
	\item In line 5 the function \textit{getSizeLayer} is used to obtain the number of units in the layer. This value will depend on the number of characteristics of the training set ($TrainData$) and the percentage applied to the corresponding layer, which is established by the \textit{PPL} parameter.
    
	\item The function \textit{getAELayer} (called in line 6 and defined in line 14) retrieves a layer of the AE model. The layer allows to obtain a new representation of the data given as first parameter ($modelData$). The number of units in the AE layer generated in this iteration will be given by the second parameter ($sizeLayer$). Firstly, the AE is initialized with the corresponding structure (line 15). The number of units in the hidden layer is given by the variable computed in the previous step, and the weights are randomly initialized. Secondly, for each train instance the following steps take place:
    
	\begin{enumerate}
		\item The AE is used to obtain the output for the given instance (line 17).
		\item The deviation of the given output with respect to the actual one is calculated (line 18).
		\item The weights of the network are updated according to the obtained error (line 19).
	\end{enumerate}
    
	Finally, the generated AE layer is returned (line 21).
    
	\item The function \textit{applyAELayer} (line 7) allows to obtain a new representation of the data given as second parameter ($modelData$). To do this, the AE layer previously generated, represented by the first parameter ($aeLayer$), is used. 
    
	\item The last step consists in adding the AE layer generated in the current iteration to the complete AE model (line 8).
\end{itemize}

During classification (lines 11-12) the function \textit{classification} is used (lines 24-35). The class for the test instances given as the first parameter ($TestData$) is predicted. The process performed internally in this function is to transform each test instance using the AE model generated in the training phase ($aeModel$), producing a new instance, more compact and representative. (line 27). This new set of features is used to predict a class with a classifier based on distances, using for each new example its \textit{k} nearest neighbors (line 28). Finally, this function returns the error rate ($result$) for the total set of test instances (line 33). As can be seen,  classification is conducted in a lower-dimensional space, mitigating the problems associated with a high number of variables.

At this point, it should be clarified that the update of weights (lines 16-20) is done using mini-batch gradient descent \cite{hinton2010practical}. This is a variation of the gradient descent algorithm that splits the training dataset into small batches that are used to calculate the model error and update the model coefficients. The reason for using this technique its better performance when dealing with large dataset.

From the previous description is can be inferred how AEkNN accomplish the objective of addressing classification with high-dimensional data. On the one hand, aiming to reduce the effects of working with a large number of variables, AEs have been used to transform such data into a lower-dimensional space. On the other hand, the classification phase is founded on the advantages of IBL. In Section \ref{Experimentation}, the performance of AEkNN is analyzed.

\section{Experimental study}\label{Experimentation}
In order to demonstrate the improvements provided by AEkNN, the algorithm proposed in the present work, an experimental study was conducted. It has been structured into three steps, all of them using the same set of datasets:

\begin{itemize}
	\item The objective of the first phase is to determine how the \textit{PPL} parameter in AEkNN influences the obtained results. For this purpose, classification results for all considered configurations are compared in subsection \ref{SelectConfiguration}. At the end, the value of the \textit{PPL} parameter that offers the best results is selected. 
    
	\item The second phase aims to verify whether AEkNN with the selected configuration improves the results provided by the classic kNN algorithm. In subsection \ref{ComparationBase}, the results of both algorithms are compared.
	
	\item The third phase of the experimentation aims to assess the competitiveness of AEkNN against traditional dimensionality reduction algorithms, in particular, principal component analysis (PCA) and linear discriminant analysis (LDA). In subsection \ref{ComparationRD}, the results of the three algorithms are compared.
\end{itemize}

Subsection \ref{Framework} describes the experimental framework, moreover the following subsections present the results and their analysis.

\subsection{Experimental framework}\label{Framework}
The conducted experimentation aims to show the benefits of AEkNN over a set of datasets with different characteristics. Their traits are shown in Table \ref{TblDatasets}. The datasets' origin is shown in the column named Ref. For all executions, datasets are given as input to the algorithms applying a 2$\times$5 folds cross validation scheme.

\begin{table*}[h]
	\centering
	%\footnotesize
    \setlength{\tabcolsep}{12pt}
	\caption{Characteristics of the datasets used in the experimentation.}
	\label{TblDatasets}
	\begin{tabular}{l | rrr | l | l}
		\toprule
		&  \multicolumn{3}{c|}{Number of} &  & \\
		\multicolumn{1}{c|}{Dataset} & Samples & Features & Classes & \multicolumn{1}{c|}{Type} & Ref \\
		\midrule
		image & 2310 & 19   & 7 &  Real  & \cite{uci} \\
		drive & 58509 & 48 & 11      &  Real & \cite{uci} \\
		coil2000 & 9822   & 85    & 2   &   Integer  & \cite{coil} \\
		dota & 102944 & 116 & 2      &  Real & \cite{uci} \\
		nomao & 1970  & 118  & 2  &  Real & \cite{uci} \\
		batch & 13910  & 128  & 6   & Real   & \cite{batch} \\
		musk & 6598  & 168   & 2      &    Integer  & \cite{uci} \\
		semeion & 1593 & 256 & 10 & Integer &  \cite{uci} \\
		madelon & 2000  & 500   & 2     &  Real  & \cite{madelon} \\
		hapt & 10929  & 561   & 12  & Real  & \cite{Hapt} \\
		isolet & 7797   & 617    & 26   &    Real  & \cite{isolet} \\
		mnist & 70000 & 784 & 10      &  Integer & \cite{Mnist} \\
		microv1 & 360   & 1300  & 10   &  Real  & \cite{uci} \\
		microv2 & 571 & 1300   & 20 & Real  & \cite{uci} \\
		\bottomrule
	\end{tabular}
\end{table*}

In both phases of experimentation, the value of \textit{k} for the classifier kNN and for AEkNN will be 5, since it is the recommended value in the related literature. In addition, to compare classification results  was necessary to compute several evaluation measures. In this experimentation, Accuracy (\ref{F1}), F-Score (\ref{F2}) and area under the ROC curve (AUC) (\ref{F5}) were used. 

Accuracy (\ref{F1}) is the proportion of true results among the total number of cases examined. 
\begin{equation}
 \textit{Accuracy} = \frac{TP + TN}{TP + TN + FP + FN} \label{F1} \end{equation}

Where \textit{TP} stands for true positives, instances correctly identified. \textit{FP} is the false positives, instances incorrectly identified. \textit{TN} represents the true negatives, instances correctly rejected. \textit{FN} corresponds to false negatives, instances incorrectly rejected.

F-Score is the harmonic mean of Precision (\ref{F3}) and Recall (\ref{F4}), considering Precision as the proportions of positive results that are true positive results and Recall as the proportion of positives that are correctly identified as such. These measures are defined by the Eqs. (\ref{F2}), (\ref{F3}), (\ref{F4}):

\begin{equation}
 \textit{F-Score} = 2 * \frac{Precision * Recall}{Precision + Recall} \label{F2}
 \end{equation}

\begin{equation}
 \textit{Precision} = \frac{TP}{TP + FP} \label{F3}
 \end{equation}

\begin{equation}
 \textit{Recall} = \frac{TP}{TP + FN} \label{F4}
 \end{equation}

Finally, AUC is the probability that a classifier will rank a randomly chosen positive instance higher than a randomly chosen negative one. AUC is given by the Eq. (\ref{F5}):

\begin{equation} \textit{AUC} = \int_{\infty}^{-\infty} \textit{TPR}(T)\textit{FPR}(T)  dT \label{F5}\end{equation}

Where \textit{TPR} stands for the true positive rate and \textit{FPR} is false positive rate.

The significance of results obtained in this experimentation is verified by appropriate statistical tests. Two different tests are used in the present study:

\begin{itemize}
	\item In the first part, the Friedman test \cite{Friedman} is used to rank the different AEkNN configurations and to establish if any statistical differences exist between them.
    
	\item In the second part, the Wilcoxon \cite{Wilcoxon} non-parametric sign rank test is used. The objective is to verify if there are significant differences between the results obtained by AEkNN and kNN.
	
\end{itemize}

These experiments were run in a cluster made up of 8 computers, having 2 CPUs (2.33 GHz) and 7 GB RAM each of them. The AEkNN algorithm and the experimentation was coded in R language \cite{R}, relying on the H2O package \cite{h2o} for some DL-related functions.

\subsection{PPL Parameter Analysis}\label{SelectConfiguration}

AEkNN has a parameter, named \textit{PPL}, that establishes the configuration of the model. This parameter allows selecting different architectures, both in number of layers (depth) and number of neurons per layer. 

The datasets used (see Table \ref{TblDatasets}) have disparate number of input features, so the architectures will be defined according to this trait. Table \ref{TblConfiguration} shows the considered configurations. For each model the number of hidden layers, as well as the number of neurons in each layer, is shown. The latter is indicated as a percentage of the number of initial characteristics. Finally, the notation of the associated \textit{PPL} parameter is provided.

\begin{table*}[h!]
	\centering
	%\footnotesize
	\addtolength{\tabcolsep}{-3pt}
    \setlength{\tabcolsep}{8pt}
	\caption{Configurations used in the experimentation and PPL parameter.}\label{TblConfiguration}
	\begin{tabular}{l | c | ccc | c}
		\toprule
		& &  \multicolumn{3}{c |}{Number of neurons (\%)} & \\
		& \# hidden layers & Layer 1 & Layer 2 & Layer 3 & PPL parameter\\
		\midrule
		AEkNN 1 & 1  & 25 & - & -  & (25)\\
		AEkNN 2 & 1   & 50 & - & - & (50)\\
		AEkNN 3 & 1  & 75 & - & - & (75)\\
		AEkNN 4 & 3 & 150 & 25 & 150 & (150, 25, 150)\\
		AEkNN 5 & 3   & 150 & 50 & 150 & (150, 50, 150)\\
		AEkNN 6 & 3  & 150 & 75 & 150 & (150, 75, 150)\\
		\bottomrule
	\end{tabular}
\end{table*}

\begin{figure*}[h!]
	\centering
	\includegraphics[width=\textwidth]{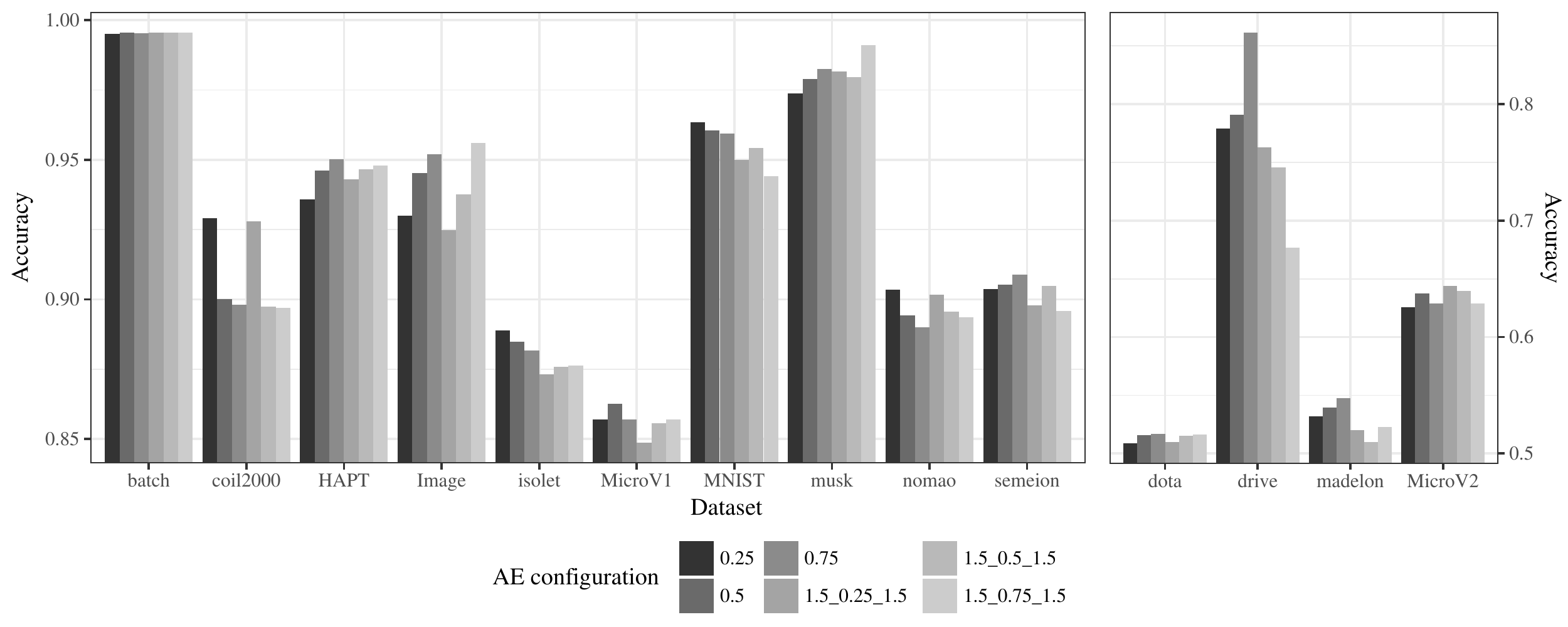}
	\caption{Accuracy classification results for test data.} \label{Accuracy_result}
\end{figure*}

\begin{figure*}[h!]
	\centering
	\includegraphics[width=\textwidth]{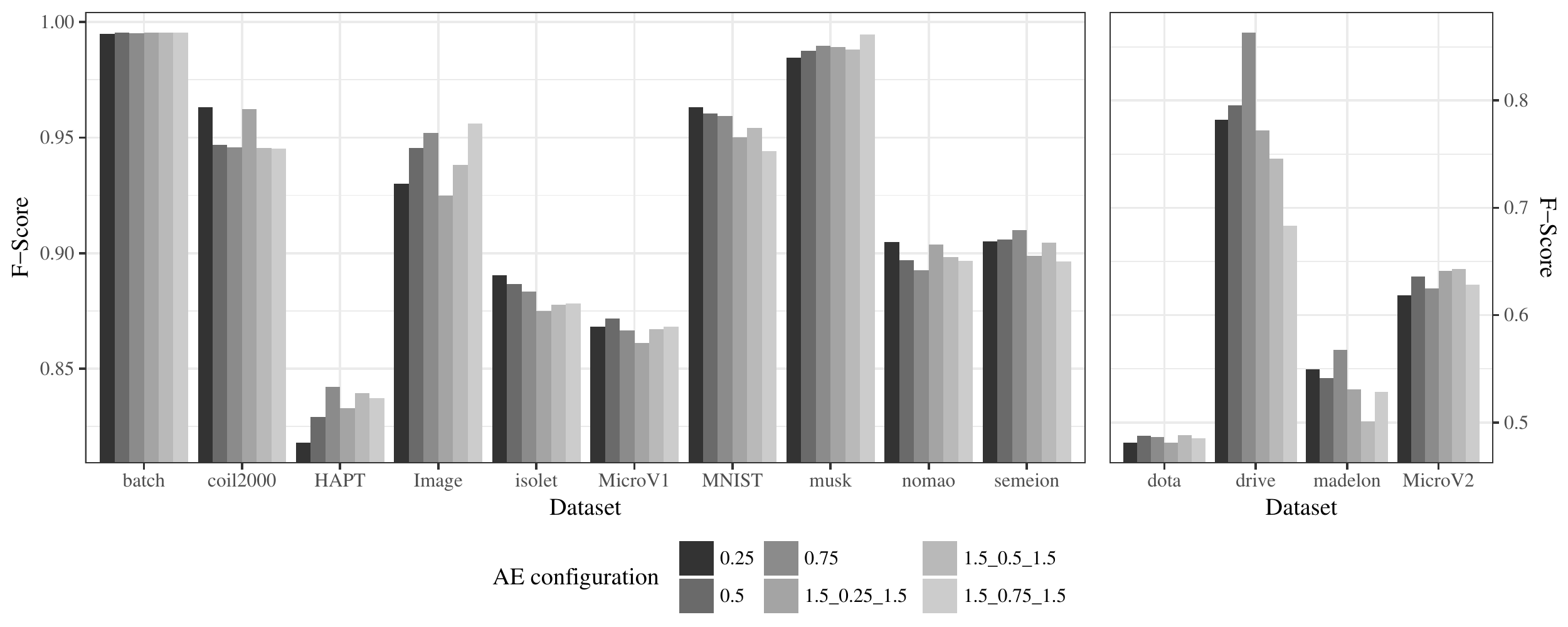}
	\caption{F-Score classification results for test data.} \label{FScore_result}
\end{figure*}

\begin{figure*}[h!]
	\centering
	\includegraphics[width=\textwidth]{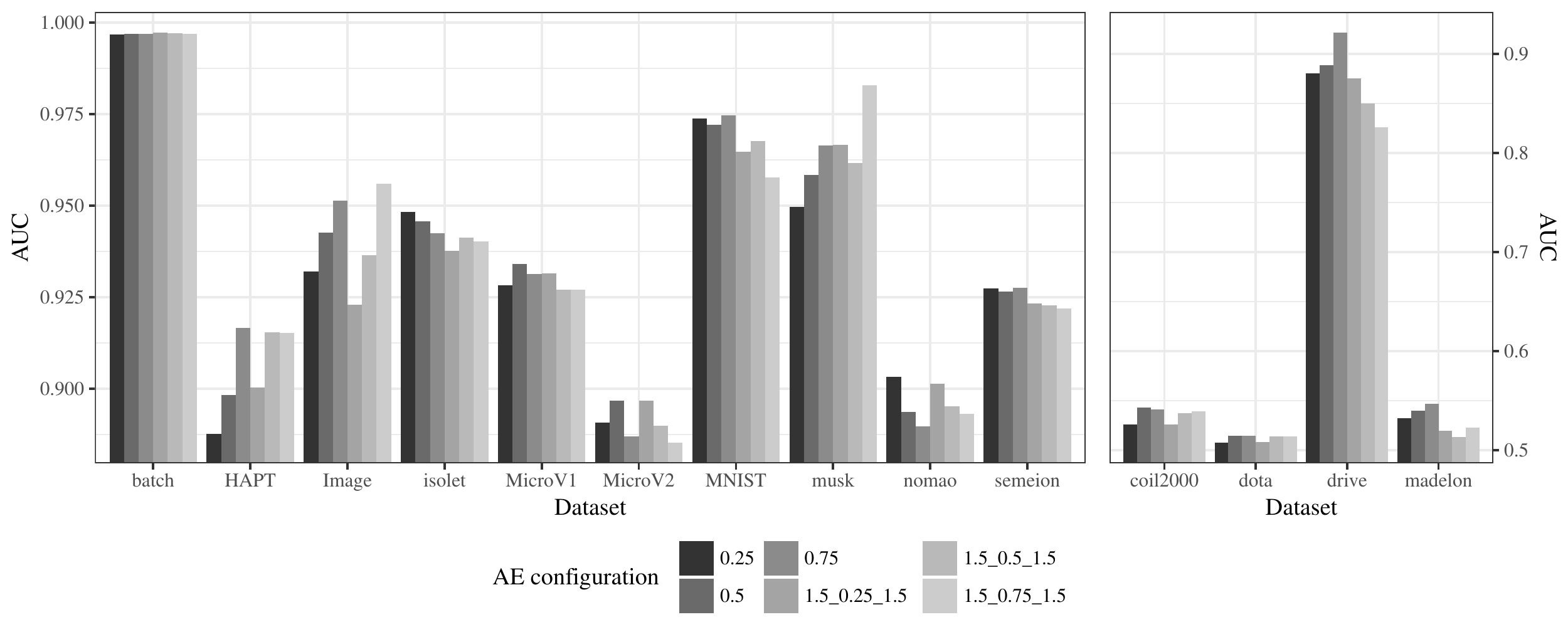}
	\caption{AUC classification results for test data.}
	\label{AUC_result}
\end{figure*}

The results produced by the different configurations considered are presented grouped by metric. Table \ref{Accuracy} shows the results for Accuracy, Table \ref{FScore} for F-Score and Table \ref{AUC} for AUC. These results are also graphically represented. Aiming to optimize the visualization, two plots with different scales have been produced for each metric. Fig. \ref{Accuracy_result} represents the results for Accuracy, Fig. \ref{FScore_result} for F-Score and Fig. \ref{AUC_result} for AUC.

\begin{table*}[h!]
	\centering
	%\footnotesize
	\setlength{\tabcolsep}{8pt}
	\addtolength{\tabcolsep}{-2pt}
	\caption{Accuracy classification results for test data.} 
	\label{Accuracy}
	\begin{tabular}{lrrrrrr}
		\toprule
		Dataset & {\scriptsize (0.25)} & {\scriptsize (0.5)} & {\scriptsize (0.75)} & {\scriptsize (1.5, 0.25, 1.5)} & {\scriptsize (1.5, 0.5, 1.5)} & {\scriptsize (1.5, 0.75, 1.5)} \\
		%{\normalsize Dataset} & & & & & &\\
		\midrule
		image & 0.930 & 0.945 & 0.952 & 0.925 & 0.938 & \textbf{0.956} \\
		drive & 0.779 & 0.791 & \textbf{0.862} & 0.763 & 0.746 & 0.677 \\
		coil2000 & \textbf{0.929} & 0.900 & 0.898 & 0.928 & 0.897 & 0.897 \\
		dota & 0.509 & 0.516 & \textbf{0.517} & 0.510 & 0.515 & 0.516 \\ 
		nomao & \textbf{0.904} & 0.894 & 0.890 & 0.902 & 0.896 & 0.894 \\ 
		batch & 0.995 & 0.995 & 0.995 & \textbf{0.996} & \textbf{0.996} & \textbf{0.996} \\
		musk & 0.974 & 0.979 & 0.983 & 0.982 & 0.980 & \textbf{0.991} \\
		semeion & 0.904 & 0.905 & \textbf{0.909} & 0.898 & 0.905 & 0.896 \\
		madelon & 0.532 & 0.540 & \textbf{0.547} & 0.520 & 0.510 & 0.523 \\   
		hapt & 0.936 & 0.946 & \textbf{0.950} & 0.943 & 0.947 & 0.948 \\
		isolet & \textbf{0.889} & 0.885 & 0.882 & 0.873 & 0.876 & 0.876 \\ 
		mnist & \textbf{0.963} & 0.960 & 0.959 & 0.950 & 0.954 & 0.944 \\
		microv1 & 0.857 & \textbf{0.863} & 0.857 & 0.849 & 0.856 & 0.857 \\ 
		microv2 & 0.625 & 0.638 & 0.629 & \textbf{0.644} & 0.639 & 0.629 \\ 	  
		\bottomrule
	\end{tabular}
\end{table*}

The results presented in Table \ref{Accuracy} and in Fig. \ref{Accuracy_result} show the Accuracy obtained for AEkNN with different \textit{PPL} values. These results indicate that there is no configuration that works best for all datasets. The configurations with three hidden layers obtain the best results in 4 out of 14 datasets, whereas the configurations with one hidden layer win in 10 out of 14 datasets. This trend can also be seen in the graphs.

\begin{table*}[h!]
	\centering
	%\footnotesize
    \setlength{\tabcolsep}{8pt}
	\addtolength{\tabcolsep}{-2pt}
	\caption{F-Score classification results for test data.} 
	\label{FScore}
	\begin{tabular}{lrrrrrr}
		\toprule
		Dataset & {\scriptsize (0.25)} & {\scriptsize (0.5)} & {\scriptsize (0.75)} & {\scriptsize (1.5, 0.25, 1.5)} & {\scriptsize (1.5, 0.5, 1.5)} & {\scriptsize (1.5, 0.75, 1.5)} \\
		%{\scriptsize Dataset} & & & & & &\\
		\midrule
		image & 0.930 & 0.945 & 0.952 & 0.925 & 0.938 & \textbf{0.956} \\ 
		drive & 0.782 & 0.796 & \textbf{0.863} & 0.772 & 0.746 & 0.683 \\
		coil2000 & \textbf{0.963} & 0.947 & 0.946 & 0.962 & 0.946 & 0.945 \\ 
		dota & 0.481 & 0.487 & 0.486 & 0.481 & \textbf{0.488} & 0.485 \\ 
		nomao & \textbf{0.905} & 0.897 & 0.893 & 0.904 & 0.898 & 0.897 \\
		batch & \textbf{0.995} & \textbf{0.995} & \textbf{0.995} & \textbf{0.995} & \textbf{0.995} & \textbf{0.995} \\
		musk & 0.984 & 0.988 & 0.990 & 0.989 & 0.988 & \textbf{0.995} \\
		semeion & 0.905 & 0.906 & \textbf{0.910} & 0.899 & 0.905 & 0.896 \\ 
		madelon & 0.549 & 0.542 & \textbf{0.567} & 0.531 & 0.501 & 0.529 \\
		hapt & 0.818 & 0.829 & \textbf{0.842} & 0.833 & 0.839 & 0.837 \\ 
		isolet & \textbf{0.890} & 0.887 & 0.883 & 0.875 & 0.878 & 0.878 \\		
		mnist & \textbf{0.963} & 0.960 & 0.959 & 0.950 & 0.954 & 0.944 \\
		microv1 & 0.868 & \textbf{0.872} & 0.867 & 0.861 & 0.867 & 0.868 \\ 
		microv2 & 0.619 & 0.636 & 0.625 & 0.641 & \textbf{0.643} & 0.628 \\
		\bottomrule
	\end{tabular}
\end{table*}

Table \ref{FScore} and Fig. \ref{FScore_result} show the F-Score obtained by AEkNN with different \textit{PPL} values. The values indicate that the configurations with one single hidden layer get better results in 11 out of 14 datasets. The configuration with \textit{PPL} = (0.25) and with \textit{PPL} = (0.75) are the ones winning more times (5). The version with \textit{PPL} = (0.25) shows disparate results, the best values for some cases and bad results for other cases, for example with \textit{hapt}, \textit{image} or \textit{microv2}. Although the version with \textit{PPL} = (0.75) wins the same number of times, its results are more balanced.

\begin{table*}[h!]
	%\centering
	%\footnotesize
	\setlength{\tabcolsep}{8pt}
	\addtolength{\tabcolsep}{-2pt}
	\caption{AUC classification results for test data.} 
	\label{AUC}
	\begin{tabular}{lrrrrrr}
		\toprule
		Dataset & {\scriptsize (0.25)} & {\scriptsize (0.5)} & {\scriptsize (0.75)} & {\scriptsize (1.5, 0.25, 1.5)} & {\scriptsize (1.5, 0.5, 1.5)} & {\scriptsize (1.5, 0.75, 1.5)} \\
		%{\scriptsize Dataset} & & & & & &\\
		\midrule
		image & 0.932 & 0.943 & 0.951 & 0.923 & 0.936 & \textbf{0.956} \\
		drive & 0.881 & 0.889 & \textbf{0.922} & 0.875 & 0.850 & 0.826 \\ 
		coil2000 & 0.526 & \textbf{0.543} & 0.541 & 0.526 & 0.538 & 0.539 \\
		dota & 0.508 & 0.514 & \textbf{0.515} & 0.508 & 0.514 & 0.514 \\ 
		nomao & \textbf{0.903} & 0.894 & 0.890 & 0.901 & 0.895 & 0.893 \\
		batch & \textbf{0.997} & \textbf{0.997} & \textbf{0.997} & \textbf{ 0.997} & \textbf{0.997} & \textbf{0.997} \\
		musk & 0.950 & 0.958 & 0.966 & 0.967 & 0.962 & \textbf{0.983} \\ 
		semeion & 0.927 & 0.927 & \textbf{0.928} & 0.923 & 0.923 & 0.922 \\
		madelon & 0.532 & 0.540 & \textbf{0.547} & 0.520 & 0.513 & 0.523 \\
		hapt & 0.888 & 0.898 & \textbf{0.917} & 0.900 & 0.916 & 0.915 \\ 
		isolet & \textbf{ 0.948} & 0.946 & 0.942 & 0.938 & 0.941 & 0.940 \\ 
		mnist & 0.974 & 0.972 & \textbf{0.975} & 0.965 & 0.968 & 0.958 \\
		microv1 & 0.928 & \textbf{0.934} & 0.931 & 0.932 & 0.927 & 0.927 \\ 
		microv2 & 0.891 & \textbf{0.897} & 0.887 & \textbf{0.897} & 0.890 & 0.885 \\
		\bottomrule
	\end{tabular}
\end{table*}

%\clearpage

In Table \ref{AUC} the results for AUC obtained with AEkNN can be seen. Fig. \ref{AUC_result} represents those results. For this metric, it can be appreciated that single hidden layer structures work better, obtaining top results in 12 out of 14 datasets. However, a configuration that works best for all cases has not been found.

Summarizing, the results presented before show the metrics obtained for AEkNN with different \textit{PPL} values. These results show some variability.  Which \textit{PPL} value is the overall best cannot be determined, since for each dataset there is a setting that works best. However, some initial trends can be drawn. Single-layer configurations get more better results than configurations with three hidden layers. Also, the configurations with \textit{PPL} = (0.75) and \textit{PPL} = (0.5) offer close to the best value in most cases, while the other configurations sometimes are far from the best value.

Once the results have been obtained, it is necessary to determine if there are statistically significant differences for each one of them in order to select the best configuration. To do this, the Friedman test \cite{Friedman} will be applied. Average ranks obtained by applying the Friedman test for Accuracy, F-Score and AUC measures are shown in Table \ref{ranking}. In addition, Table \ref{friedman} shows the different \textit{p-values} obtained by the Friedman test.

\begin{table*}[h!]
	\centering
	%\footnotesize
	\setlength{\tabcolsep}{8pt}
	\addtolength{\tabcolsep}{-2pt}
	\caption{Average rankings of the different PPL values by measure}
	\label{ranking}
	\begin{tabular}{cc |cc | cc}
		\toprule
		\multicolumn{2}{c |}{Accuracy}&  \multicolumn{2}{c |}{F-Score} & \multicolumn{2}{c}{AUC}\\
		PPL&Ranking&PPL&Ranking&PPL&Ranking\\
		\midrule
		(0.75) & 2.679 & (0.5) & 2.923 & (0.75) & 2.357\\
		(0.5) & 2.857 & (0.75) & 3.000 & (0.5) & 2.786\\
		(0.25) & 3.714 & (0.25) & 3.429 & (1.5, 0.25, 1.5) & 3.786\\
		(1.5, 0.75, 1.5) & 3.821 & (1.5, 0.5, 1.5) & 3.500 & (0.25) & 3.857\\
		(1.5, 0.5, 1.5) & 3.892 & (1.5, 0.25, 1.5) & 4.071 & (1.5, 0.5, 1.5) & 4.000\\
		(1.5, 0.25, 1.5) & 4.036 & (1.5, 0.75, 1.5) & 4.071 & (1.5, 0.75, 1.5) & 4.214\\
		\bottomrule
	\end{tabular}
\end{table*}

\begin{table}[h!]
	\centering
	%\footnotesize
	\addtolength{\tabcolsep}{-2pt}
    \setlength{\tabcolsep}{12pt}
	\caption{Results of Friedman's test (\textit{p-values})}
	\label{friedman}
	\begin{tabular}{c|c|c}
		\toprule
		Accuracy&F-Score&AUC\\
		\midrule
		0.236 &  0.423 & 0.049  \\
		\bottomrule
	\end{tabular}
\end{table}

As can be observed in Table \ref{friedman}, for AUC (wich is considerer a stronger performance metric) there are statistically significant differences between the considered \textit{PPL} values if we set the \textit{p-value} threshold to the usual range [0.05,~0.1]. However, for Accuracy and F-Score there are no statistically significant differences. In addition, in the rankings obtained, it can be seen that there are two specific configurations that offer better results than the remaining ones. In the three rankings presented, the results with \textit{PPL} = (0.75) and \textit{PPL} = (0.5) appear first, clearly highlighted with respect to the other values. Therefore, it is considered that these two configurations are the best ones. Thus, the results of AEkNN with both configurations will be compared against the kNN algorithm.

\subsection{AEkNN vs kNN}\label{ComparationBase}

This second part is focused on determining if the results obtained with the proposed algorithm, AEkNN, improve those obtained with the kNN algorithm. To do so, a comparison will be made between the results obtained with AEkNN, using the values of the \textit{PPL} parameter selected in the previous section, and the results obtained with kNN algorithm on the same datasets.

\begin{figure*}[h!]
	\centering
	\includegraphics[width=\textwidth]{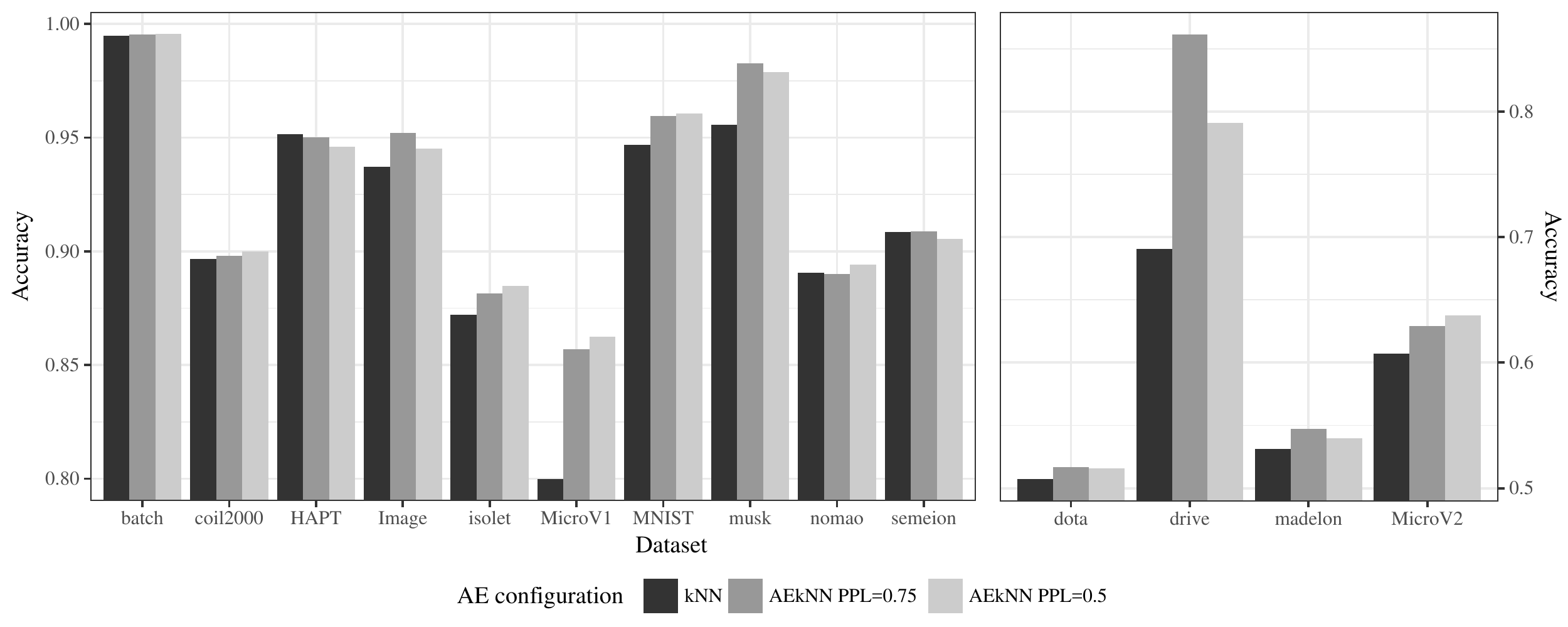}
	\caption{Accuracy results for test data.} 
	\label{Accuracy2_result}
\end{figure*}

\begin{figure*}[h!]
	\centering
	\includegraphics[width=\textwidth]{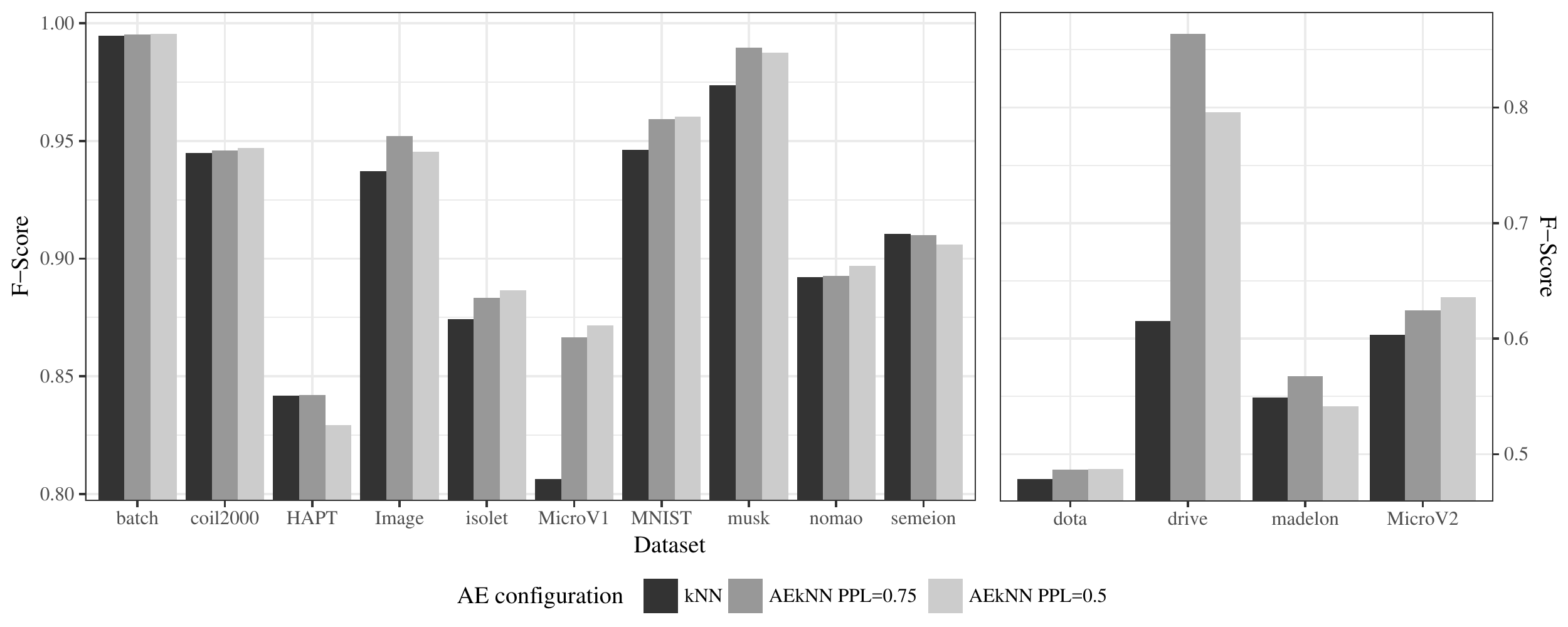}
	\caption{F-Score results for test data.} \label{FScore2_result}
\end{figure*}

\begin{figure*}[h!]
	\centering
	\includegraphics[width=\textwidth]{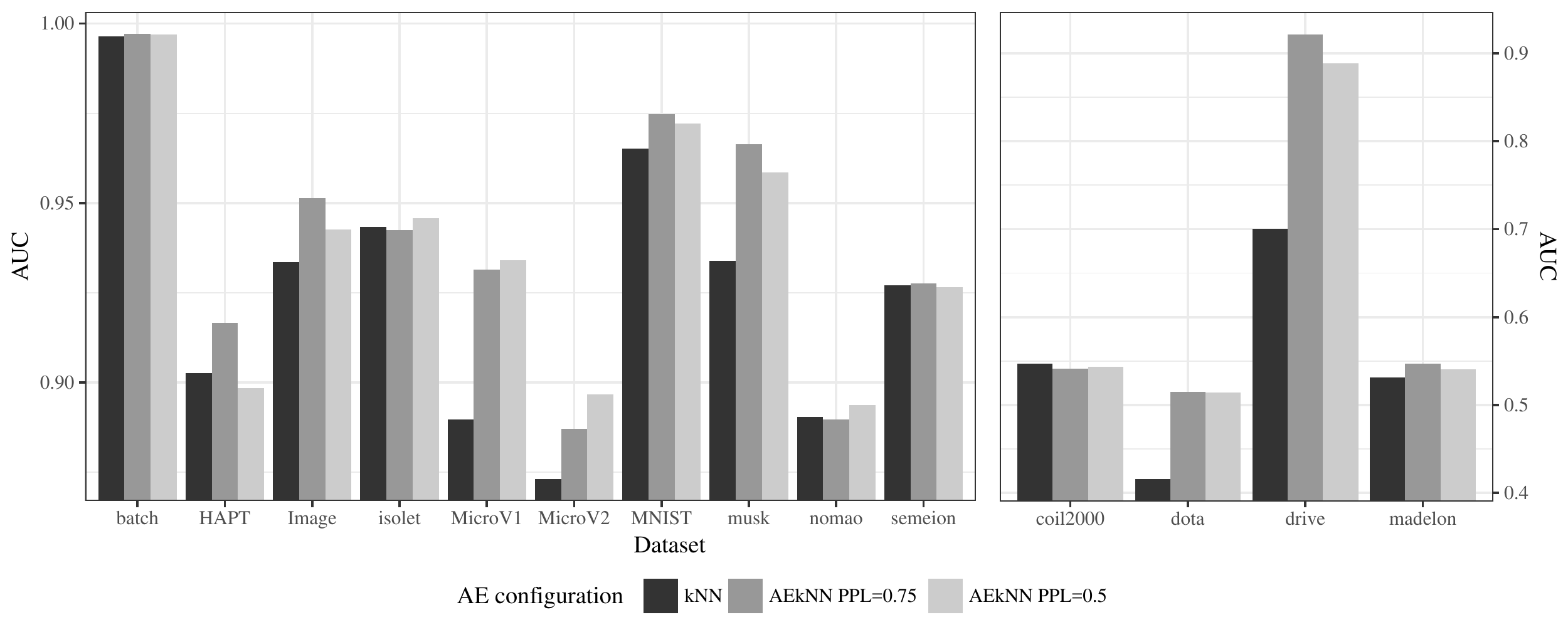}
	\caption{AUC results for test data.} 
	\label{AUC2_result}
\end{figure*}

\begin{figure*}[h!]
	\centering
	\includegraphics[width=\textwidth]{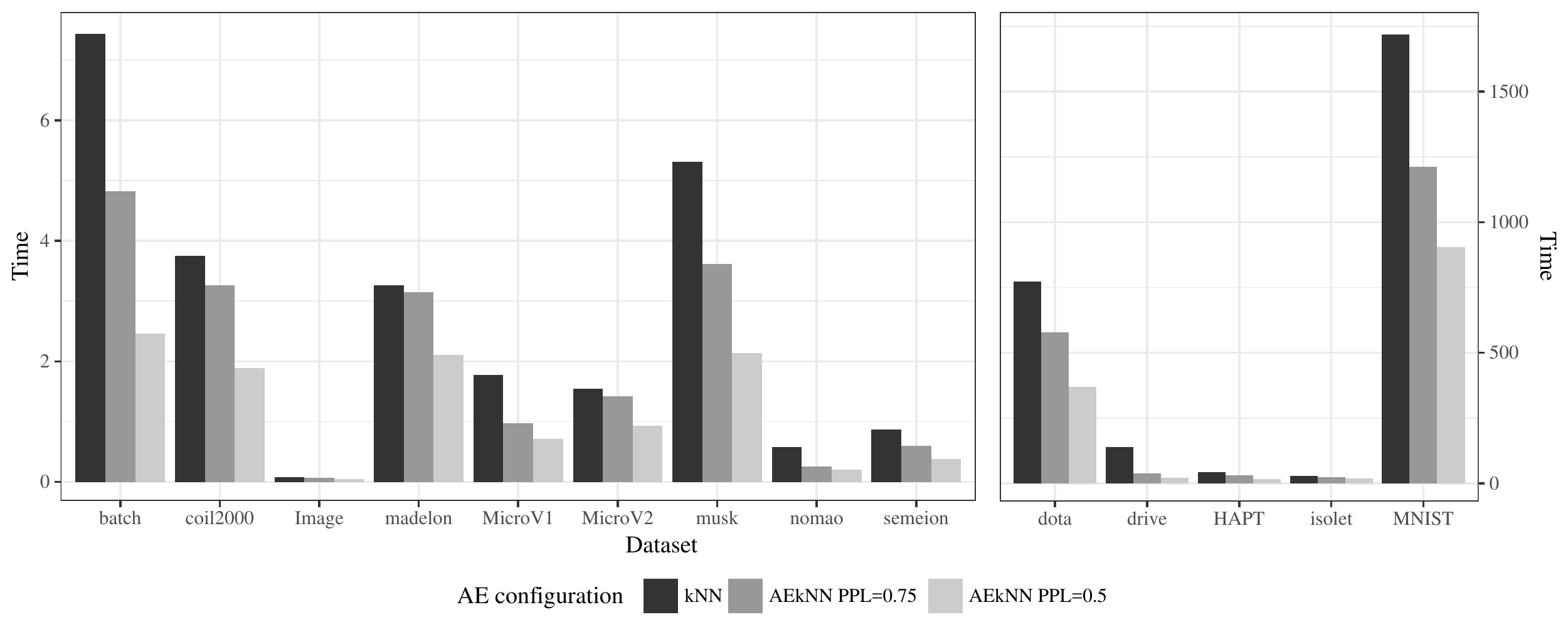}
	\caption{Time results for test data.} \label{Time_result}
\end{figure*}

First, Table \ref{tblComparationBase} shows the results for each one of the datasets and considered measures, including running time. The results for both algorithms are presented jointly, and the best ones are highlighted in bold. Two plots have been generated for each metric aiming to optimize data visualization, as in the previous phase, since the range of results was very broad. Fig. \ref{Accuracy2_result} represents the results for Accuracy, Fig. \ref{FScore2_result} for F-Score, Fig. \ref{AUC2_result} for AUC, and Fig. \ref{Time_result} for runtime.

%\clearpage
\begin{table*}[h!]
	\footnotesize
	\centering
	\caption{Classification results of AEkNN (with different PPL) and kNN algorithm for test data} 
	\setlength{\tabcolsep}{12pt}
	\addtolength{\tabcolsep}{-10pt}
	\label{tblComparationBase}
	\begin{tabular}{l|rrr|rrr|rrr|rrr}
		\toprule
		& \multicolumn{3}{c|}{Accuracy} & \multicolumn{3}{c|}{F-Score} & \multicolumn{3}{c|}{AUC} & \multicolumn{3}{c}{Time (seconds)}\\
		& kNN & \multicolumn{2}{c|}{AEkNN} & kNN & \multicolumn{2}{c|}{AEkNN} & kNN & \multicolumn{2}{c|}{AEkNN} &kNN & \multicolumn{2}{c}{AEkNN} \\ 
		Dataset &  & (0.75) & (0.5) &  & (0.75) & (0.5) &  & (0.75) & (0.5) &  & (0.75) & (0.5) \\ 
		\midrule
		image & 0.937 & \textbf{0.952} & 0.945 & 0.937 & \textbf{0.952} & 0.945 & 0.934 & \textbf{0.951} & 0.943 & 0.074 & 0.073 & \textbf{0.052} \\
		drive & 0.691 & \textbf{0.862} & 0.791 & 0.615 & \textbf{0.863} & 0.796 & 0.700 & \textbf{0.922} & 0.889 & 139.623 & 36.977 & \textbf{20.479} \\
		coil2000 & 0.897 & 0.898 & \textbf{0.900} & 0.945 & 0.946 & \textbf{0.947} & \textbf{0.547} & 0.541 & 0.543 & 3.753 & 3.262 & \textbf{1.886} \\
		dota & 0.507 & \textbf{0.517} & 0.516 & 0.479 & 0.486 & \textbf{0.487} & 0.416 & \textbf{0.515} & 0.514 & 772.219 & 578.437 & \textbf{370.599} \\ 
		nomao & 0.891 & 0.890 & \textbf{0.894} & 0.892 & 0.893 & \textbf{0.897} & 0.891 & 0.890 & \textbf{0.894} & 0.582 & 0.252 & \textbf{0.200} \\
		batch & \textbf{0.995} & \textbf{0.995} & \textbf{0.995} & \textbf{0.995} & \textbf{0.995} & \textbf{0.995} & 0.996 &\textbf{ 0.997} & \textbf{0.997} & 7.433 & 4.829 & \textbf{2.459} \\
		musk & 0.956 & \textbf{0.983} & 0.979 & 0.974 & \textbf{0.990 }& 0.988 & 0.934 & \textbf{0.966} & 0.958 & 5.317 & 3.613 & \textbf{2.144} \\ 
		semeion & 0.908 & \textbf{0.909} & 0.905 & \textbf{0.910} & \textbf{0.910} & 0.906 & 0.927 & \textbf{0.928} & 0.927 & 0.865 & 0.601 & \textbf{0.381} \\
		madelon & 0.531 & \textbf{0.547} & 0.540 & 0.549 & \textbf{0.567} & 0.542 & 0.532 & \textbf{0.547} & 0.540 & 3.267 & 3.150 & \textbf{2.109} \\
		hapt & \textbf{0.951} & 0.950 & 0.946 & \textbf{0.842} & \textbf{0.842} & 0.829 & 0.903 & \textbf{0.917} & 0.898 & 41.673 & 30.625 & \textbf{16.365} \\ 	 
		isolet & 0.872 & 0.882 & \textbf{0.885} & 0.874 & 0.883 & \textbf{0.887} & 0.943 & 0.942 & \textbf{0.946} & 27.563 & 24.748 & \textbf{19.538} \\ 
		mnist & 0.947 & 0.959 & \textbf{0.960} & 0.946 & 0.959 & \textbf{0.960} & 0.965 & \textbf{0.975} & 0.972 & 1720.547 & 1213.168 & \textbf{904.223} \\ 
		microv1 & 0.800 & 0.857 & \textbf{0.863} & 0.806 & 0.867 & \textbf{0.872} & 0.890 & 0.931 &\textbf{ 0.934} & 1.776 & 0.977 & \textbf{0.709} \\ 
		microv2 & 0.607 & 0.629 & \textbf{0.638} & 0.603 & 0.625 & \textbf{0.636} & 0.873 & 0.887 & \textbf{0.897} & 1.542 & 1.425 & \textbf{0.933} \\  
		\bottomrule
	\end{tabular}
\end{table*}

The results shown in Table \ref{tblComparationBase} indicate that AEkNN works better than kNN for most datasets considering Accuracy. On the one hand, the version of AEkNN with \textit{PPL} = (0.75) improves kNN in 11 out of 14 cases, obtaining the best overall results in 6 of them. On the other hand, the version of AEkNN with \textit{PPL} = (0.5) obtains better results than kNN in 11 out of 14 cases, being the best configuration in 6 of them. In addition, kNN only obtains one best result. Fig. \ref{Accuracy2_result} confirms this trend. It can be observed that the right bars, where AEkNN results are represented, are higher in most datasets.

%\clearpage

Analyzing the data corresponding to the metric F-Score, presented in Table \ref{tblComparationBase}, it can be observed that AEkNN produces an overall improvement over kNN. The AEkNN version with \textit{PPL} = (0.75) improves kNN in 11 out of 14 cases, obtaining the best overall results in 5 of them. The version of AEkNN with \textit{PPL} = (0.5) obtains better results than kNN in 10 out of 14 cases, being the best configuration in 7 of them. kNN does not obtain any best result. How the results of the versions corresponding to AEkNN produce higher values than those corresponding to kNN can be seen in Fig. \ref{FScore2_result}.

The data related to AUC, presented in Table \ref{tblComparationBase}, also show that AEkNN works better than kNN. In this case, the two versions of AEkNN improve kNN in 11 out of 14 cases each one, obtaining the best results in 13 out of 14 cases. Fig. \ref{AUC2_result} shows that the trend is increasing towards the versions of the new algorithm. kNN only gets a best result, specifically with the \textit{coil2000} dataset, maybe due to the low number of features in this dataset. Nonetheless, AEkNN gets better than kNN in the rest of metrics for \textit{coil2000} dataset. 

%\clearpage

The running times for both algorithms are presented in Table \ref{tblComparationBase} and Fig. \ref{Time_result}. As can be seen, the configuration that takes less time to classify is the one corresponding to AEkNN with \textit{PPL} = (0.5), obtaining the lowest value for all datasets. This is due to the higher compression of the data achieved by this configuration. In the same way, AEkNN with \textit{PPL} = (0.75) obtains better results than the algorithm kNN in all cases. 

Summarizing, it can be observed that the results obtained through AEkNN improve those obtained with the original kNN algorithm for most of the datasets. AEkNN, despite the transformation of the input space to reduce dimensionality, the quality of the results in terms of classification performance are better than those of kNN in most cases. In addition, in terms of classification time, it can be noted how AEkNN with higher compression of information significantly reduces the time spent on classification, without having a negative impact on the other measures.

To determine if there are statistically significant differences between the obtained results, the proper statistical test has been conducted. For this purpose, the Wilcoxon test will be performed, comparing each version of AEkNN against the results of the classical kNN algorithm. In Table \ref{wilcoxon}, the results obtained for Wilcoxon tests are shown.

\begin{table*}[h]
	\centering
	\caption{Result of Wilcoxon's test (\textit{p-values}) comparing kNN vs AEkNN}
	\label{wilcoxon}
	\begin{tabular}{c|c|c|c|c}\hline
		&Accuracy&F-Score&AUC&Time\\\hline
		AEkNN with PPL=(0.75) & 0.0017 & 0.0003 & 0.0085 & 0.0002 \\
		AEkNN with PPL=(0.5) & 0.0023 & 0.0245 & 0.0107 & 0.0001\\
		\hline
	\end{tabular}
\end{table*}

As can be seen the \textit{p-values} are rather low, so that statistical significant differences between the two AEkNN versions and the original kNN algorithm in all considered measures exist can be concluded, considering the \textit{p-value} threshold within the usual [0.05,~0.1] range. On the one hand, taking into account Accuracy, F-Score and AUC, the configuration with best results is that performing a 75\% of feature reduction. Therefore, it is the optimal solution from the point of view of predictive performance. The reason for this might be that there is less compression of the data, therefore, there is less loss of information compared to the other considered configuration  (50\%). On the other hand, considering running time, the configuration with best results is that performing a 50\% of feature reduction. It is not surprising that having less features allows to compute distances in less time.

\subsection{AEkNN vs PCA / LDA}\label{ComparationRD}

The objective of this third part is to assess the competitiveness of AEkNN against traditional dimensionality reduction algorithms. In particular, the algorithms used will be PCA \cite{Pearson1901,Hotelling1933} and LDA \cite{fukunaga2013introduction,yu2001direct}, since they are traditional algorithms that offer good results in this task \cite{martinez2001pca}. To do so, a comparison will be made between the results obtained with AEkNN, using the values of the \textit{PPL} parameter selected in section \ref{ComparationBase}, and the results obtained with PCA and LDA algorithm on the same datasets. It is important to note that the number of features selected with these methods will be the same as with the AEkNN algorithm, so there are two executions for each algorithm.

\begin{figure*}[h!]
	\centering
	\includegraphics[width=\textwidth]{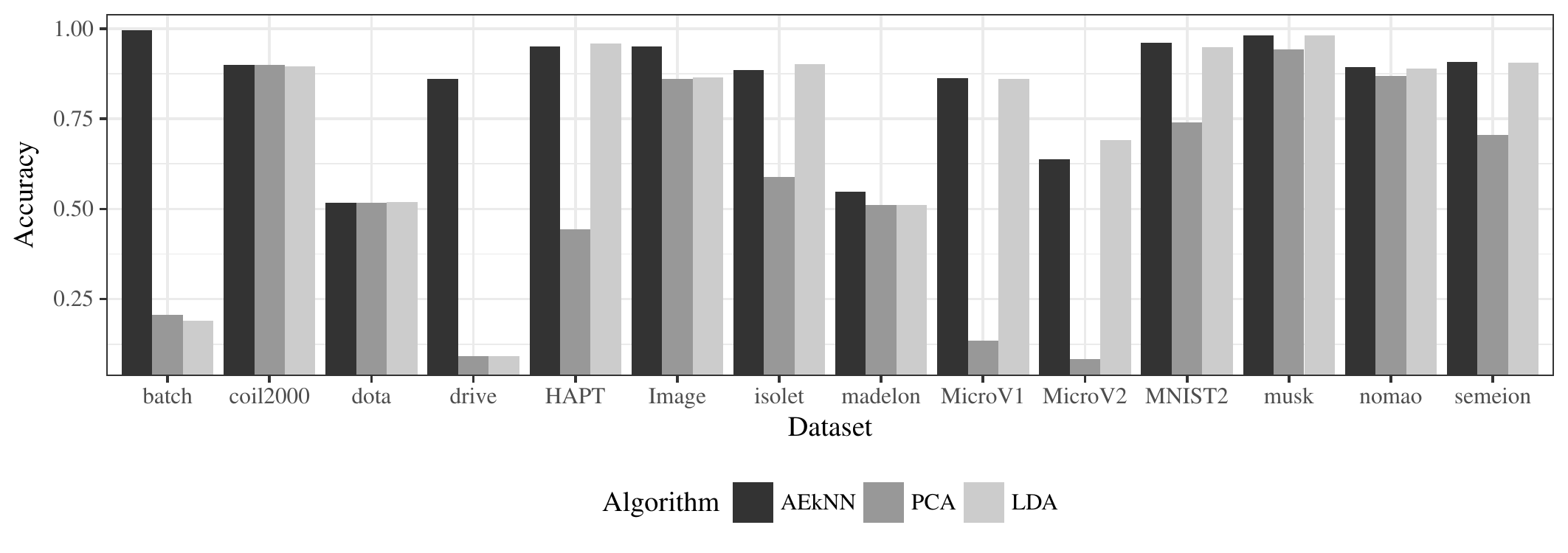}
	\caption{Accuracy results for test data.} \label{Accuracy_resultDR2}
\end{figure*}

\begin{figure*}[h!]
	\centering
	\includegraphics[width=\textwidth]{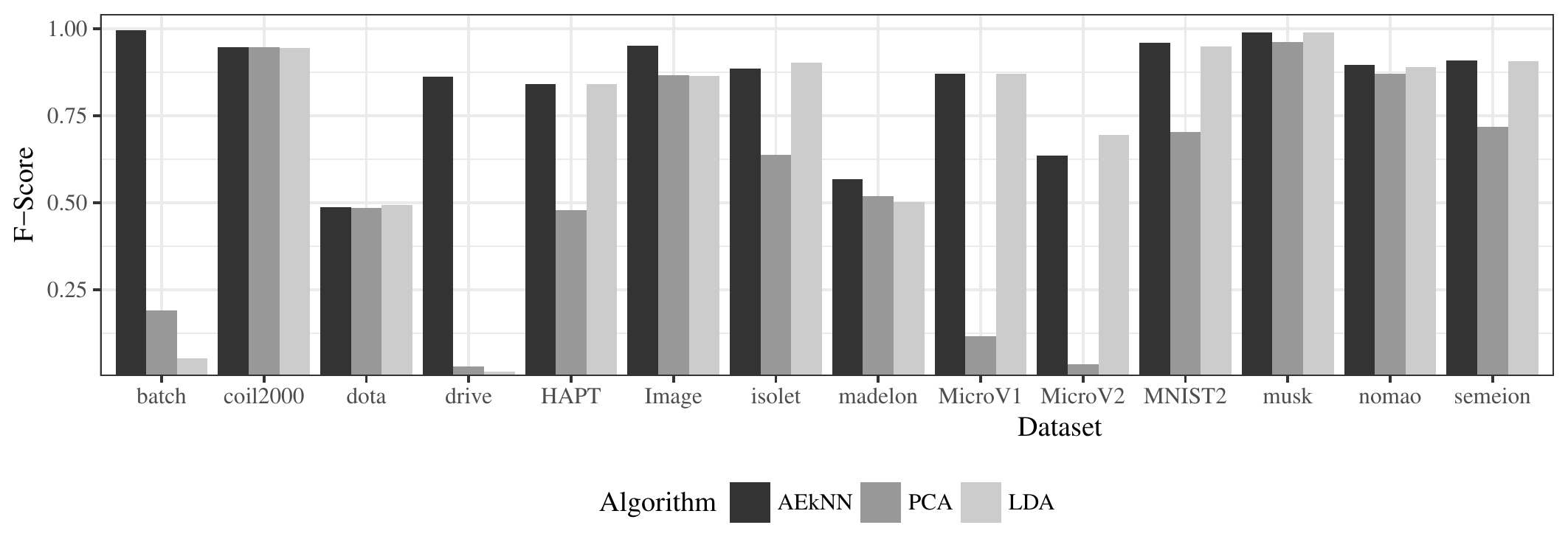}
	\caption{F-Score results for test data.} \label{FScore_resultDR2}
\end{figure*}

\begin{figure*}[h!]
	\centering
	\includegraphics[width=\textwidth]{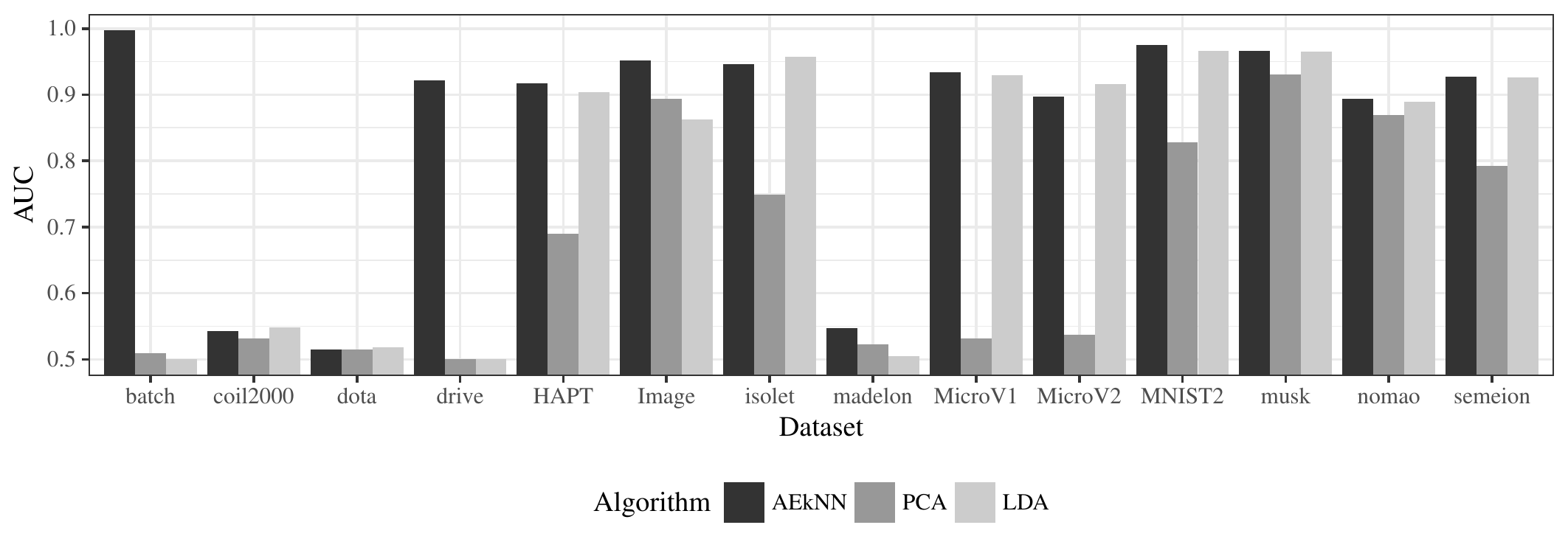}
	\caption{AUC results for test data.} \label{AUC_resultDR2}
\end{figure*}

First, Table \ref{tblComparationRD} shows the results for each one of the datasets and considered measures. The results for the three algorithms are presented jointly, and the best ones are highlighted in bold. One plot have been generated for each metric aiming to optimize data visualization. In this case, the graphs represent the best value of the two configurations for each algorithm in order to better visualize the differences between the three methods. Fig. \ref{Accuracy_resultDR2} represents the results for Accuracy, Fig. \ref{FScore_resultDR2} for F-Score and Fig. \ref{AUC_resultDR2} for AUC.

The results shown in Table \ref{tblComparationRD} indicate that AEkNN works better than PCA and LDA for most datasets considering Accuracy. On the one hand, the version of AEkNN with \textit{PPL} = (0.75) improves PCA in in 12 out of 14 cases and LDA in 9 out of 14 cases, obtaining the best overall results in 7 of them. On the other hand, the version of AEkNN with \textit{PPL} = (0.5) obtains better results than PCA in 13 out of 14 cases and LDA in 8 out of 14 cases, being the best configuration in 4 of them. In addition, LDA only obtains the best result in 4 cases. Fig. \ref{Accuracy_resultDR2} confirms this trend. It can be observed that the bars, where AEkNN results are represented, are higher in most datasets.

Analyzing the data corresponding to the metric F-Score, presented in Table \ref{tblComparationRD}, it can be observed that AEkNN produces an overall improvement over PCA and LDA. The AEkNN version with \textit{PPL} = (0.75) improves PCA in 12 out of 14 cases and LDA in 10 out of 14 cases, obtaining the best overall results in 6 of them. The version of AEkNN with \textit{PPL} = (0.5) obtains better results than PCA in all cases and LDA in 8 out of 14 cases, being the best configuration in 4 of them. PCA does not obtain any best result and LDA obtains the best result in 4 cases. How the results of the versions corresponding to AEkNN show values higher than those corresponding to kNN can be seen in Fig. \ref{FScore_resultDR2}.

\vspace*{-0.4cm}
The data related to AUC, presented in Table \ref{tblComparationRD}, also show that AEkNN works better than PCA and LDA. The AEkNN version with \textit{PPL} = (0.75) improves PCA in 13 out of 14 cases and LDA in 10 out of 14 cases, obtaining the best overall results in 7 of them. The version of AEkNN with \textit{PPL} = (0.5) obtains better results than PCA in 13 out of 14 cases and LDA in 8 out of 14 cases, being the best configuration in 4 of them. LDA only gets the best result in 4 cases. Fig. \ref{AUC_resultDR2} confirms this trend.

\begin{landscape}
\mbox{}\vfill
%\hspace*{-2cm}
\begin{minipage}[b]{1\linewidth}
	\centering
    \scriptsize
    \setlength{\tabcolsep}{6pt}
    \addtolength{\tabcolsep}{-3pt}
    \captionof{table}{Accuracy, F-Score and AUC classification results of AEkNN (with different PPL), LDA and PCA for test data}
    \label{tblComparationRD}
	\begin{tabular}{l|rrrrrr|rrrrrr|rrrrrr}
    	\toprule
		& \multicolumn{6}{c|}{Accuracy} & \multicolumn{6}{c|}{F-Score} & \multicolumn{6}{c}{AUC} \\
		& \multicolumn{2}{c}{AEkNN} & \multicolumn{2}{c}{PCA} & \multicolumn{2}{c|}{LDA} & \multicolumn{2}{c}{AEkNN} & \multicolumn{2}{c}{PCA} & \multicolumn{2}{c|}{LDA} & \multicolumn{2}{c}{AEkNN} & \multicolumn{2}{c}{PCA} & \multicolumn{2}{c}{LDA}\\ 
		Dataset & (0.75) & (0.5) & (0.75) & (0.5) & (0.75) & (0.5) & (0.75) & (0.5) & (0.75) & (0.5) & (0.75) & (0.5) & (0.75) & (0.5) & (0.75) & (0.5) & (0.75) & (0.5) \\ 
		\midrule
		image & \textbf{0.952} & 0.945 & 0.563 & 0.862 & 0.676 & 0.866 & \textbf{0.952} & 0.945 & 0.580 & 0.867 & 0.672 & 0.866 & \textbf{0.951} & 0.943 & 0.734 & 0.893 & 0.780 & 0.862\\
        drive & \textbf{0.862} & 0.791 & 0.091 & 0.092 & 0.091 & 0.091 & \textbf{0.863} & 0.796 & 0.023 & 0.030 & 0.015 & 0.015 & \textbf{0.922} & 0.889 & 0.500 & 0.491 & 0.500 & 0.500\\
		coil2000 & 0.898 & \textbf{0.900} & 0.825 & 0.899 & 0.895 & 0.895 & 0.946 & \textbf{0.947} & 0.861 & 0.946 & 0.944 & 0.944 & 0.541 & 0.543 & 0.523 & 0.532 & 0.545 & \textbf{0.549}\\ 
		dota & 0.517 & 0.516 & 0.514 & 0.517 & 0.519 & \textbf{0.520} & 0.486 & 0.487 & 0.486 & 0.486 & 0.489 & \textbf{0.493} & 0.515 & 0.514 & 0.513 & 0.515 & 0.517 & \textbf{0.519}\\ 
		nomao & 0.890 & \textbf{0.894} & 0.798 & 0.869 & 0.818 & 0.889 & 0.893 & \textbf{0.897} & 0.821 & 0.871 & 0.851 & 0.891 & 0.890 & \textbf{0.894} & 0.797 & 0.869 & 0.817 & 0.889 \\ 
		batch & \textbf{0.995} & \textbf{0.995} & 0.195 & 0.206 & 0.189 & 0.152 & \textbf{0.995} & \textbf{0.995} & 0.176 & 0.190 & 0.053 & 0.044 & \textbf{0.997} & \textbf{0.997} & 0.502 & 0.510 & 0.500 & 0.500\\
		musk & \textbf{0.983} & 0.979 & 0.942 & 0.931 & 0.982 & 0.982 & \textbf{0.990} & 0.988 & 0.962 & 0.945 & 0.989 & 0.989 & \textbf{0.966} & 0.958 & 0.912 & 0.930 & 0.965 & 0.965\\ 		
		semeion & \textbf{0.909} & 0.905 & 0.598 & 0.706 & 0.907 & 0.890 & \textbf{0.910} & 0.906 & 0.624 & 0.718 & 0.906 & 0.893 & \textbf{0.928} & 0.927 & 0.732 & 0.792 & 0.926 & 0.913 \\
		madelon & \textbf{0.547} & 0.540 & 0.506 & 0.511 & 0.501 & 0.512 & \textbf{0.567} & 0.542 & 0.519 & 0.503 & 0.503 & 0.491 & \textbf{0.547} & 0.540 & 0.523 & 0.510 & 0.505 & 0.505 \\
		hapt & 0.950 & 0.946 & 0.179 & 0.443 & 0.959 & \textbf{0.960} & \textbf{0.842} & 0.829 & 0.180 & 0.478 & 0.841 & 0.841 & \textbf{0.917} & 0.898 & 0.553 & 0.690 & 0.903 & 0.899 \\ 		 
		isolet & 0.882 & 0.885 & 0.424 & 0.590 & 0.886 & \textbf{0.902} & 0.883 & 0.887 & 0.504 & 0.638 & 0.888 & \textbf{0.903} & 0.942 & 0.946 & 0.660 & 0.748 & 0.951 & \textbf{0.957}\\ 
		mnist & 0.959 & \textbf{0.960} & 0.740 & 0.640 & 0.949 & 0.943 & 0.959 & \textbf{0.960} & 0.704 & 0.704 & 0.949 & 0.943 & 0.975 & \textbf{0.972} & 0.828 & 0.798 & 0.966 & 0.954 \\ 
		microv1 & 0.857 & \textbf{0.863} & 0.135 & 0.135 & 0.818 & 0.861 & 0.867 & \textbf{0.872} & 0.116 & 0.116 & 0.823 & 0.871 & 0.931 & \textbf{0.934} & 0.532 & 0.532 & 0.898 & 0.929\\ 
		microv2 & 0.629 & 0.638 & 0.058 & 0.082 & 0.649 & \textbf{0.690} & 0.625 & 0.636 & 0.005 & 0.037 & 0.644 & \textbf{0.695} & 0.887 & 0.897 & 0.500 & 0.537 & 0.891 & \textbf{0.916} \\ 
	\bottomrule
	\end{tabular}
\end{minipage}
\vfill
\end{landscape}

Summarizing, it can be observed that the results obtained through AEkNN improve those obtained with the PCA and LDA algorithms for most of the datasets. The quality of the results with AEkNN in terms of classification performance are better than those of PCA and LDA in most cases. This means that the high-level features obtained by the AEkNN algorithm provide more relevant information than those obtained by the PCA and LDA algorithms. Therefore, the results obtained by classifying with the AEkNN algorithm improve those obtained with PCA and LDA. 

Previously, the data obtained in the experimentation have been presented and a comparison between them is made. However, it is necessary to verify if there are significant differences between the data corresponding to the different algorithms. To do this, the Friedman test \cite{Friedman} will be applied. Average ranks obtained by applying the Friedman test for Accuracy, F-Score and AUC measures are shown in Table \ref{rankingRD}. In addition, Table \ref{friedmanRD} shows the different \textit{p-values} obtained by the Friedman test.

\begin{table*}[h!]
	\centering
	%\footnotesize
	\setlength{\tabcolsep}{4pt}
	\addtolength{\tabcolsep}{-2pt}
	\caption{Average rankings of the different PPL values by measure}
	\label{rankingRD}
	\begin{tabular}{cc |cc | cc}
		\toprule
		\multicolumn{2}{c |}{Accuracy}&  \multicolumn{2}{c |}{F-Score} & \multicolumn{2}{c}{AUC}\\
		Algorithm&Ranking&Algorithm&Ranking&Algorithm&Ranking\\
		\midrule
		AEkNN PPL=0.75 & 2.286 & AEkNN PPL=0.75 & 2.143 & AEkNN PPL=0.75 & 1.929\\ 
		AEkNN PPL=0.5 & 2.357 & AEkNN PPL=0.5 & 2.214 & AEkNN PPL=0.5 & 2.500\\ 
		LDA PPL=0.5 & 3.000 & LDA PPL=0.75 & 3.429 & LDA PPL=0.5 & 2.929\\ 
		LDA PPL=0.75 & 3.571 & LDA PPL=0.5 & 3.429 & LDA PPL=0.75 & 3.500\\ 
		PCA PPL=0.5 & 4.321 & PCA PPL=0.5 & 4.429 & PCA PPL=0.5 & 4.679\\ 
		PCA PPL=0.75 & 5.464 &PCA PPL=0.75 & 5.357 & PCA PPL=0.75 & 5.464\\ 
		\bottomrule
	\end{tabular}
\end{table*}

\begin{table}[h!]
	\centering
	%\footnotesize
	\addtolength{\tabcolsep}{-2pt}
	\setlength{\tabcolsep}{12pt}
	\caption{Results of Friedman's test (\textit{p-values})}
	\label{friedmanRD}
	\begin{tabular}{c|c|c}
		\toprule
		Accuracy&F-Score&AUC\\
		\midrule
		1.266e-05 &  7.832e-06 & 7.913e-07  \\
		\bottomrule
	\end{tabular}
\end{table}

As can be observed in Table \ref{friedmanRD}, for Accuracy, F-Score and AUC there are statistically significant differences between the different \textit{PPL} values if we set the \textit{p-value} threshold to the usual range [0.05,~0.1]. It can be seen that AEkNN with \textit{PPL} = (0.75) offer better results than the remaining ones. In the three rankings presented, the AEkNN configurations with \textit{PPL} = (0.75) and \textit{PPL} = (0.5) appear first, clearly highlighted with respect to the other values. Therefore, it is considered that AEkNN obtains better predictive performance, since the reduction of dimensionality generates more significant features.

\subsection{General Guidelines on the use of AEkNN}\label{GeneralGuidelines}

AEkNN could be considered as a robust algorithm on the basis of the previous analysis. The experimental work demonstrates that it has good performance with the two \textit{PPL} considered values. From the conducted experimentation some guidelines can also be extracted:

\begin{itemize}
	\item When working with very high-dimensional datasets, it is recommended to use AEkNN with the \textit{PPL} = (0.5) configuration. In this study, this configuration has obtained the best results for datasets having more than 600 features. The reason is that the input data has a larger number of features and allows a greater reduction without losing relevant information. Therefore, AEkNN can compress more in these cases. 
	
	\item When using binary datasets with a lower dimensionality, the AEkNN algorithm with the \textit{PPL} = (0.5) configuration continues to be the best choice. In our experience, this configuration has shown to work better for binary datasets with a number of features around 100. In these cases, the compression may be higher since it is easier to discriminate by class.
	
	\item For all other datasets, the choice of configuration for AEkNN depends on the indicator to be enhanced. On the one hand, if the goal is to achieve the best possible predictive performance, the configuration with \textit{PPL} = (0.75) must be chosen. In these cases, AEkNN needs to use more original information. On the other hand, when the interest is to optimize the running time, while maintaining improvements in predictive performance with respect to kNN, the configuration with \textit{PPL} = (0.5) is the best selection. The reason is in the higher compression of the data. AEkNN needs less time to classify lower-dimensional data.
\end{itemize}

Summarizing, the configuration of AEkNN must be adapted to the data traits to obtain optimal results. For this, a series of tips have been established.

\section{Concluding remarks}\label{Concluding}

In this paper, a new classification algorithm called AEkNN has been proposed. This algorithm is based on kNN but aims to mitigate the problem that arises while working with high-dimensional data. To do so, AEkNN internally incorporates a model-building phase aimed to perform a reduction of the feature space, using AEs for this purpose. The main reason that has led to the design of AEkNN are the good results that have been obtained by AEs when they are used to generate higher-level features. AEkNN relies on an AE  to extract a reduced representation of a higher level that replaces the original data.

In order to determine if the proposed algorithm offers better results than the kNN algorithm, a experimentation has been carried out. Firstly, the analysis of these results have allowed to determine which AE structure works better. 
Furthermore, in the second part of the conducted experimentation, the results of the best configurations have been compared with the results produced by kNN. As has been stated, the results of AEkNN improve those obtained by kNN for all the metrics. In addition, AEkNN offers a considerable improvement with respect to the time invested in the classification.

In addition, a comparison has been made with other traditional methods applied to this problem, in order to verify that the AEkNN algorithm improves the results when carrying out the dimensionality reduction task. For this, AEkNN has been compared with LDA and PCA. The results show that the proposed AEkNN algorithm improves the performance in classification for most of the dataset used. This occurs because the features generated with the proposed algorithm are more significant and provide more relevant information to the classification using distance-based algorithms.

In conclusion, AEkNN is able to reduce the adverse effects of high-dimensional data while performing instance-based classification, improving both running time and classification performance.  These results show that the use of AEs can be helpful to solve this kind of obstacle, opening up new possibilities of future work in which they are applied to help solve similar problems presented by other traditional models.

\section*{Acknowledgment}
The work of F. Pulgar was supported by the Spanish Ministry of Education under the FPU National Program (Ref. FPU16/00324). This work was partially supported by the Spanish Ministry of Science and Technology under project TIN2015-68454-R.

\bibliographystyle{spbasic}
%\bibliography{DeepLearning}

\begin{thebibliography}{94}
	\providecommand{\natexlab}[1]{#1}
	\providecommand{\url}[1]{{#1}}
	\providecommand{\urlprefix}{URL }
	\expandafter\ifx\csname urlstyle\endcsname\relax
	\providecommand{\doi}[1]{DOI~\discretionary{}{}{}#1}\else
	\providecommand{\doi}{DOI~\discretionary{}{}{}\begingroup
		\urlstyle{rm}\Url}\fi
	\providecommand{\eprint}[2][]{\url{#2}}
	
	\bibitem[{Aha et~al(1991)Aha, Kibler, and Albert}]{aha1991}
	Aha DW, Kibler D, Albert MK (1991) {Instance-Based Learning Algorithms}.
	Machine Learning 6(1):37--66, \doi{10.1023/A:1022689900470}
	
	\bibitem[{Aiello et~al(2016)Aiello, Kraljevic, Maj, and with contributions
		from~the H2O.ai~team}]{h2o}
	Aiello S, Kraljevic T, Maj P, with contributions from~the H2Oai~team (2016)
	h2o: R Interface for H2O.
	\urlprefix\url{https://CRAN.R-project.org/package=h2o}, r package version
	3.8.1.3
	
	\bibitem[{Alpaydin(2014)}]{Alpa2014}
	Alpaydin E (2014) Introduction to machine learning. MIT press
	
	\bibitem[{Altman(1992)}]{altman}
	Altman NS (1992) {An Introduction to Kernel and Nearest-Neighbor Nonparametric
		Regression}. The American Statistician 46(3):175--185,
	\doi{10.1080/00031305.1992.10475879}
	
	\bibitem[{Atkeson et~al(1997)Atkeson, Moorey, Schaalz, Moore, and
		Schaal}]{Atkeson1997}
	Atkeson CG, Moorey AW, Schaalz S, Moore AW, Schaal S (1997) {Locally Weighted
		Learning}. Artificial Intelligence 11:11--73, \doi{10.1023/A:1006559212014}
	
	\bibitem[{Bache and Lichman(2013)}]{uci}
	Bache K, Lichman M (2013) {UCI Machine Learning Repository}.
	\urlprefix\url{http://www.ics.uci.edu/$\sim$mlearn/MLRepository.html}
	
	\bibitem[{Belkin and Niyogi(2003)}]{belkin2003laplacian}
	Belkin M, Niyogi P (2003) Laplacian eigenmaps for dimensionality reduction and
	data representation. Neural Computation 15(6):1373--1396,
	\doi{10.1162/089976603321780317}
	
	\bibitem[{Bellman(1957)}]{Richard1957}
	Bellman R (1957) {Dynamic Programming}. Princeton University Press
	
	\bibitem[{Bellman(1961)}]{Richard1961}
	Bellman R (1961) {Adaptive control processes: A guided tour}. Princeton
	University Press
	
	\bibitem[{Bengio(2009)}]{bengio2009}
	Bengio Y (2009) {Learning Deep Architectures for AI}. Foundations and Trends in
	Machine Learning 2(1):1--127, \doi{10.1561/2200000006}
	
	\bibitem[{Bengio(2013)}]{Bengio2013}
	Bengio Y (2013) {Deep Learning of Representations: Looking Forward}. In:
	International Conference on Statistical Language and Speech Processing, pp
	1--37, \doi{10.1007/978-3-642-39593-2\_1}
	
	\bibitem[{Bengio et~al(2013)Bengio, Courville, and Vincent}]{bengio2013_1}
	Bengio Y, Courville A, Vincent P (2013) {Representation Learning: A Review and
		New Perspectives}. Pattern Analysis and Machine Intelligence, IEEE
	Transactions on 35(8):1798--1828, \doi{10.1109/TPAMI.2013.50}
	
	\bibitem[{Beyer et~al(1999)Beyer, Goldstein, Ramakrishnan, and
		Shaft}]{Beyer1999}
	Beyer K, Goldstein J, Ramakrishnan R, Shaft U (1999) {When Is ''Nearest
		Neighbor'' Meaningful?} In: International conference on database theory, pp
	217--235, \doi{10.1007/3-540-49257-7\_15}
	
	\bibitem[{Bhatia and Rani(2018)}]{Bhatia2018}
	Bhatia V, Rani R (2018) Dfuzzy: a deep learning-based fuzzy clustering model
	for large graphs. Knowledge and Information Systems
	\doi{10.1007/s10115-018-1156-3},
	\urlprefix\url{https://doi.org/10.1007/s10115-018-1156-3}
	
	\bibitem[{Burges(2005)}]{burges2005}
	Burges CJC (2005) {Geometric Methods for Feature Extraction and Dimensional
		Reduction}, Springer US, pp 59--91. \doi{10.1007/0-387-25465-X\_4}
	
	\bibitem[{Casillas et~al(2001)Casillas, Cord{\'o}n, Del~Jesus, and
		Herrera}]{casillas2001genetic}
	Casillas J, Cord{\'o}n O, Del~Jesus MJ, Herrera F (2001) Genetic feature
	selection in a fuzzy rule-based classification system learning process for
	high-dimensional problems. Information Sciences 136(1):135--157,
	\doi{10.1016/S0020-0255(01)00147-5}
	
	\bibitem[{Charte et~al(2018)Charte, Charte, García, del Jesus, and
		Herrera}]{CHARTE201878}
	Charte D, Charte F, García S, del Jesus MJ, Herrera F (2018) A practical
	tutorial on autoencoders for nonlinear feature fusion: Taxonomy, models,
	software and guidelines. Information Fusion 44:78--96,
	\doi{10.1016/j.inffus.2017.12.007}
	
	\bibitem[{Chung et~al(2015)Chung, G{\"u}l{\c{c}}ehre, Cho, and
		Bengio}]{chung2015}
	Chung J, G{\"u}l{\c{c}}ehre C, Cho K, Bengio Y (2015) Gated feedback recurrent
	neural networks. In: Proceedings of the International Conference on Machine
	Learning, pp 2067--2075
	
	\bibitem[{Ciresan et~al(2012{\natexlab{a}})Ciresan, Giusti, Gambardella, and
		Schmidhuber}]{image2}
	Ciresan D, Giusti A, Gambardella L, Schmidhuber J (2012{\natexlab{a}}) {Deep
		Neural Networks Segment Neuronal Membranes in Electron Microscopy Images}.
	In: Advances in neural information processing systems, pp 2843--2851
	
	\bibitem[{Ciresan et~al(2012{\natexlab{b}})Ciresan, Meier, and
		Schmidhuber}]{image1}
	Ciresan D, Meier U, Schmidhuber J (2012{\natexlab{b}}) {Multi-column deep
		neural networks for image classification}. In: Proceedings of the IEEE
	Conference on Computer Vision and Pattern Recognition, pp 3642--3649,
	\doi{10.1109/CVPR.2012.6248110}
	
	\bibitem[{Coifman et~al(2005)Coifman, Lafon, Lee, Maggioni, Nadler, Warner, and
		Zucker}]{coifman2005}
	Coifman RR, Lafon S, Lee AB, Maggioni M, Nadler B, Warner F, Zucker SW (2005)
	{Geometric diffusions as a tool for harmonic analysis and structure
		definition of data: Diffusion maps}. Proceedings of the National Academy of
	Sciences 102(21):7426--7431, \doi{10.1073/pnas.0500334102}
	
	\bibitem[{Cole and Fanty(1990)}]{isolet}
	Cole R, Fanty M (1990) {Spoken letter recognition}. In: Proceedings of the
	workshop on Speech and Natural Language, pp 385--390,
	\doi{10.3115/116580.116725}
	
	\bibitem[{Cover and Hart(1967)}]{cover1967}
	Cover T, Hart P (1967) {Nearest neighbor pattern classification}. IEEE
	Transactions on Information Theory 13(1):21--27,
	\doi{10.1109/TIT.1967.1053964}
	
	\bibitem[{Dasarathy(1991)}]{dasarathy1991}
	Dasarathy BV (1991) {Nearest neighbor norms: NN pattern classification
		techniques}. IEEE Computer Society Press
	
	\bibitem[{Deng(2014)}]{Deng2014}
	Deng L (2014) {Deep Learning: Methods and Applications}. Foundations and Trends
	in Signal Processing 7(3-4):197--387, \doi{10.1561/2000000039}
	
	\bibitem[{Deng and Yu(2011)}]{deng2011}
	Deng L, Yu D (2011) {Deep convex net: A scalable architecture for speech
		pattern classification}. In: Proceedings of the Annual Conference of the
	International Speech Communication Association, pp 2285--2288
	
	\bibitem[{Deng et~al(2017)Deng, Ren, Kong, Bao, and Dai}]{deng2017hierarchical}
	Deng Y, Ren Z, Kong Y, Bao F, Dai Q (2017) A hierarchical fused fuzzy deep
	neural network for data classification. IEEE Transactions on Fuzzy Systems
	25(4):1006--1012, \doi{10.1109/TFUZZ.2016.2574915}
	
	\bibitem[{Deng et~al(2016)Deng, Zhu, Cheng, Zong, and
		Zhang}]{deng2016efficient}
	Deng Z, Zhu X, Cheng D, Zong M, Zhang S (2016) Efficient knn classification
	algorithm for big data. Neurocomputing 195:143--148,
	\doi{10.1016/j.neucom.2015.08.112}
	
	\bibitem[{Duda et~al(1973)Duda, Hart, and Stork}]{MachineLearning}
	Duda RO, Hart PE, Stork DG (1973) {Pattern Classification}. Wiley New York
	
	\bibitem[{Esteva et~al(2017)Esteva, Kuprel, Novoa, Ko, Swetter, Blau, and
		Thrun}]{esteva2017dermatologist}
	Esteva A, Kuprel B, Novoa RA, Ko J, Swetter SM, Blau HM, Thrun S (2017)
	Dermatologist-level classification of skin cancer with deep neural networks.
	Nature 542(7639):115--118, \doi{10.1038/nature21056}
	
	\bibitem[{Friedman(1937)}]{Friedman}
	Friedman M (1937) {The use of ranks to avoid the assumption of normality
		implicit in the analysis of variance}. Journal of the American Statistical
	Association 32(200):675--701, \doi{10.1080/01621459.1937.10503522}
	
	\bibitem[{Fukunaga(2013)}]{fukunaga2013introduction}
	Fukunaga K (2013) Introduction to statistical pattern recognition. Academic
	Press
	
	\bibitem[{Goldberg and Holland(1988)}]{Goldberg1988}
	Goldberg DE, Holland JH (1988) {Genetic Algorithms and Machine Learning}.
	Machine Learning 3(2):95--99, \doi{10.1023/A:1022602019183}
	
	\bibitem[{Goodfellow et~al(2016)Goodfellow, Bengio, and
		Courville}]{goodfellow2016}
	Goodfellow I, Bengio Y, Courville A (2016) {Deep Learning}. The MIT Press
	
	\bibitem[{Guyon et~al(2004)Guyon, Gunn, Ben-Hur, and Dror}]{madelon}
	Guyon I, Gunn S, Ben-Hur A, Dror G (2004) {Result Analysis of the NIPS 2003
		Feature Selection Challenge}. In: Proceedings of Neural Information
	Processing Systems, vol~4, pp 545--552
	
	\bibitem[{Harsanyi and Chang(1994)}]{harsanyi1994hyperspectral}
	Harsanyi JC, Chang CI (1994) Hyperspectral image classification and
	dimensionality reduction: An orthogonal subspace projection approach. IEEE
	Transactions on Geoscience and Remote Sensing 32(4):779--785,
	\doi{10.1109/36.298007}
	
	\bibitem[{Hinneburg et~al(2000)Hinneburg, Aggarwal, and Keim}]{hinne2000}
	Hinneburg A, Aggarwal CC, Keim DA (2000) {What Is the Nearest Neighbor in High
		Dimensional Spaces?} In: Proceedings of the International Conference on Very
	Large Databases, pp 506--515
	
	\bibitem[{Hinton(2010)}]{hinton2010practical}
	Hinton G (2010) A practical guide to training restricted boltzmann machines.
	Momentum 9(1):926, \doi{10.1007/978-3-642-35289-8\_32}
	
	\bibitem[{Hinton and Salakhutdinov(2012)}]{hinton2012}
	Hinton G, Salakhutdinov R (2012) {A better way to pretrain deep Boltzmann
		machines}. Advances in Neural Information Processing Systems (3):2447--2455
	
	\bibitem[{Hinton et~al(2012)Hinton, Deng, Yu, Dahl, Mohamed, Jaitly, Senior,
		Vanhoucke, Nguyen, and Sainath}]{speech2}
	Hinton G, Deng L, Yu D, Dahl GE, Mohamed AR, Jaitly N, Senior A, Vanhoucke V,
	Nguyen P, Sainath TN (2012) Deep neural networks for acoustic modeling in
	speech recognition: The shared views of four research groups. IEEE Signal
	Processing Magazine 29(6):82--97
	
	\bibitem[{Hinton and Salakhutdinov(2006)}]{hinton2006}
	Hinton GE, Salakhutdinov RR (2006) {Reducing the Dimensionality of Data with
		Neural Networks}. Science 313(5786):504--507, \doi{10.1126/science.1127647}
	
	\bibitem[{Hochreiter and Schmidhuber(1997)}]{hoch1997}
	Hochreiter S, Schmidhuber J (1997) {Long Short-Term Memory}. Neural Computation
	9(8):1735--1780, \doi{10.1162/neco.1997.9.8.1735}
	
	\bibitem[{Hotelling(1933)}]{Hotelling1933}
	Hotelling H (1933) {Analysis of a complex of statistical variables into
		principal components.} Journal of Educational Psychology 24(6):417--441,
	\doi{10.1037/h0071325}
	
	\bibitem[{Hughes(1968)}]{hughes}
	Hughes GF (1968) {On the Mean Accuracy of Statistical Pattern Recognizers}.
	IEEE Transactions on Information Theory 14(1):55--63,
	\doi{10.1109/TIT.1968.1054102}
	
	\bibitem[{Keyvanrad and Homayounpour(2017)}]{Keyvanrad2017}
	Keyvanrad MA, Homayounpour MM (2017) Effective sparsity control in deep belief
	networks using normal regularization term. Knowledge and Information Systems
	53(2):533--550, \doi{10.1007/s10115-017-1049-x},
	\urlprefix\url{https://doi.org/10.1007/s10115-017-1049-x}
	
	\bibitem[{Kohavi and Provost(1998)}]{kohavi1998}
	Kohavi R, Provost F (1998) {Glossary of Terms}. Machine Learning
	30(2-3):271--274, \doi{10.1023/A:1017181826899}
	
	\bibitem[{Kotsiantis(2007)}]{kotsiantis}
	Kotsiantis SB (2007) {Supervised machine learning: A review of classification
		techniques}. Informatica 31:249--268
	
	\bibitem[{Kouiroukidis and Evangelidis(2011)}]{kou2011}
	Kouiroukidis N, Evangelidis G (2011) {The Effects of Dimensionality Curse in
		High Dimensional kNN Search}. In: Proceedings of the 15th Panhellenic
	Conference on Informatics, IEEE, pp 41--45, \doi{10.1109/PCI.2011.45}
	
	\bibitem[{Krizhevsky et~al(2012)Krizhevsky, Sutskever, and {Geoffrey
			E.}}]{image3}
	Krizhevsky A, Sutskever I, {Geoffrey E} H (2012) {ImageNet Classification with
		Deep Convolutional Neural Networks}. In: Advances in neural information
	processing systems, pp 1097--1105
	
	\bibitem[{LeCun and Bengio(1995)}]{LeCun1995}
	LeCun Y, Bengio Y (1995) {Convolutional networks for images, speech, and time
		series}. The handbook of brain theory and neural networks 3361:255--258,
	\doi{10.1109/IJCNN.2004.1381049}
	
	\bibitem[{Lecun et~al(1998)Lecun, Bottou, Bengio, and Haffner}]{Mnist}
	Lecun Y, Bottou L, Bengio Y, Haffner P (1998) {Gradient-based learning applied
		to document recognition}. Proceedings of the IEEE 86(11):2278--2324,
	\doi{10.1109/5.726791}
	
	\bibitem[{LeCun et~al(2015)LeCun, Bengio, and Hinton}]{LeCun2015}
	LeCun Y, Bengio Y, Hinton G (2015) {Deep learning}. Nature 521(7553):436--444,
	\doi{10.1038/nature14539}
	
	\bibitem[{Lee et~al(2009)Lee, Grosse, Ranganath, and Ng}]{Lee2009}
	Lee H, Grosse R, Ranganath R, Ng AY (2009) {Convolutional deep belief networks
		for scalable unsupervised learning of hierarchical representations}. In:
	Proceedings of the 26th Annual International Conference on Machine Learning,
	ACM Press, vol 2008, pp 609--616, \doi{10.1145/1553374.1553453}
	
	\bibitem[{Lee and Verleysen(2007)}]{Lee2007}
	Lee JA, Verleysen M (2007) {Nonlinear Dimensionality Reduction}. Springer
	Science \& Business Media
	
	\bibitem[{Lin et~al(2010)Lin, Zhang, Zhu, and Yu}]{lin2010deep}
	Lin Y, Zhang T, Zhu S, Yu K (2010) Deep coding network. In: Advances in Neural
	Information Processing Systems, pp 1405--1413
	
	\bibitem[{Liou et~al(2014)Liou, Cheng, Liou, and Liou}]{liou2014}
	Liou CY, Cheng WC, Liou JW, Liou DR (2014) {Autoencoder for words}.
	Neurocomputing 139:84--96, \doi{10.1016/j.neucom.2013.09.055}
	
	\bibitem[{Liu and Motoda(1998)}]{liu1998}
	Liu H, Motoda H (1998) Feature extraction, construction and selection: A data
	mining perspective, vol 453. Springer Science \& Business Media,
	\doi{10.1007/978-1-4615-5725-8}
	
	\bibitem[{Liu and Motoda(2007)}]{liu2007}
	Liu H, Motoda H (2007) Computational methods of feature selection. CRC Press
	
	\bibitem[{Maillo et~al(2017{\natexlab{a}})Maillo, Luengo, Garc{\'\i}a, Herrera,
		and Triguero}]{maillo2017exact}
	Maillo J, Luengo J, Garc{\'\i}a S, Herrera F, Triguero I (2017{\natexlab{a}})
	Exact fuzzy k-nearest neighbor classification for big datasets. In: IEEE
	International Conference on Fuzzy Systems, IEEE, pp 1--6,
	\doi{10.1109/FUZZ-IEEE.2017.8015686}
	
	\bibitem[{Maillo et~al(2017{\natexlab{b}})Maillo, Ram{\'\i}rez, Triguero, and
		Herrera}]{maillo2017knn}
	Maillo J, Ram{\'\i}rez S, Triguero I, Herrera F (2017{\natexlab{b}}) knn-is: An
	iterative spark-based design of the k-nearest neighbors classifier for big
	data. Knowledge-Based Systems 117:3--15, \doi{10.1016/j.knosys.2016.06.012}
	
	\bibitem[{Mart{\'\i}nez and Kak(2001)}]{martinez2001pca}
	Mart{\'\i}nez AM, Kak AC (2001) {PCA} versus {LDA}. IEEE Transactions on
	Pattern Analysis and Machine Intelligence 23(2):228--233,
	\doi{10.1109/34.908974}
	
	\bibitem[{Min et~al(2009)Min, Stanley, Yuan, Bonner, and Zhang}]{Min2009}
	Min R, Stanley DA, Yuan Z, Bonner A, Zhang Z (2009) {A Deep Non-linear Feature
		Mapping for Large-Margin kNN Classification}. In: Proceedings of the
	International Conference on Data Mining, IEEE, pp 357--366,
	\doi{10.1109/ICDM.2009.27}
	
	\bibitem[{Pearson(1901)}]{Pearson1901}
	Pearson K (1901) {LIII. On lines and planes of closest fit to systems of points
		in space}. The London, Edinburgh, and Dublin Philosophical Magazine and
	Journal of Science 2(11):559--572, \doi{10.1080/14786440109462720}
	
	\bibitem[{{R Core Team}(2016)}]{R}
	{R Core Team} (2016) R: A Language and Environment for Statistical Computing. R
	Foundation for Statistical Computing, Vienna, Austria,
	\urlprefix\url{https://www.R-project.org/}
	
	\bibitem[{Radovanovi{\'{c}} et~al(2010)Radovanovi{\'{c}}, Nanopoulos, and
		Ivanovi{\'{c}}}]{Rado2010}
	Radovanovi{\'{c}} M, Nanopoulos A, Ivanovi{\'{c}} M (2010) {Hubs in Space :
		Popular Nearest Neighbors in High-Dimensional Data}. Journal of Machine
	Learning Research 11:2487--2531
	
	\bibitem[{Reyes-Ortiz et~al(2016)Reyes-Ortiz, Oneto, Sam{\`{a}}, Parra, and
		Anguita}]{Hapt}
	Reyes-Ortiz JL, Oneto L, Sam{\`{a}} A, Parra X, Anguita D (2016)
	{Transition-Aware Human Activity Recognition Using Smartphones}.
	Neurocomputing 171:754--767, \doi{10.1016/j.neucom.2015.07.085}
	
	\bibitem[{Rifai and Muller(2011)}]{Rifai2011}
	Rifai S, Muller X (2011) {Contractive Auto-Encoders: Explicit Invariance During
		Feature Extraction}. In: Proceedings of the 28th International Conference on
	Machine Learning, vol~85, pp 833--840
	
	\bibitem[{Roweis and Saul(2000)}]{roweis2000nonlinear}
	Roweis ST, Saul LK (2000) Nonlinear dimensionality reduction by locally linear
	embedding. Science 290(5500):2323--2326, \doi{10.1126/science.290.5500.2323}
	
	\bibitem[{Sak et~al(2014)Sak, Senior, and Beaufays}]{sak2014}
	Sak H, Senior A, Beaufays F (2014) {Long Short-Term Memory Recurrent Neural
		Network Architectures for Large Scale Acoustic Modeling}. In: Proceedings of
	the Annual Conference of the International Speech Communication Association,
	September, pp 338--342
	
	\bibitem[{Salakhutdinov et~al(2007)Salakhutdinov, Mnih, and
		Hinton}]{salakhutdinov2007restricted}
	Salakhutdinov R, Mnih A, Hinton G (2007) Restricted boltzmann machines for
	collaborative filtering. In: Proceedings of the 24th International Conference
	on Machine learning, ACM, pp 791--798, \doi{10.1145/1273496.1273596}
	
	\bibitem[{Saul et~al(2006)Saul, Weinberger, Ham, Sha, and Lee}]{saul2006}
	Saul LK, Weinberger KQ, Ham JH, Sha F, Lee DD (2006) {Spectral Methods for
		Dimensionality Reduction}. The MIT Press, pp 292--308,
	\doi{10.7551/mitpress/9780262033589.003.0016}
	
	\bibitem[{Sheskin(2004)}]{Wilcoxon}
	Sheskin DJ (2004) {Handbook of parametric and nonparametric statistical
		procedures}. Technometrics 46:1193, \doi{10.1198/tech.2004.s209}
	
	\bibitem[{Snoek et~al(2012)Snoek, Adams, and
		Larochelle}]{snoek2012nonparametric}
	Snoek J, Adams RP, Larochelle H (2012) Nonparametric guidance of autoencoder
	representations using label information. Journal of Machine Learning Research
	13(Sep):2567--2588
	
	\bibitem[{Sordoni et~al(2015)Sordoni, Bengio, Vahabi, Lioma, Grue~Simonsen, and
		Nie}]{sordoni2015hierarchical}
	Sordoni A, Bengio Y, Vahabi H, Lioma C, Grue~Simonsen J, Nie JY (2015) A
	hierarchical recurrent encoder-decoder for generative context-aware query
	suggestion. In: Proceedings of the 24th ACM International on Conference on
	Information and Knowledge Management, ACM, pp 553--562
	
	\bibitem[{Spearman(1904)}]{Spearman1904}
	Spearman C (1904) {"General Intelligence," Objectively Determined and
		Measured}. The American Journal of Psychology (2):201--292,
	\doi{10.2307/1412107}
	
	\bibitem[{Tang et~al(2017)Tang, Zhang, Wang, Feng, Roli, and Liu}]{Tang2017}
	Tang P, Zhang J, Wang X, Feng B, Roli F, Liu W (2017) Learning extremely shared
	middle-level image representation for scene classification. Knowledge and
	Information Systems 52(2):509--530, \doi{10.1007/s10115-016-1015-z},
	\urlprefix\url{https://doi.org/10.1007/s10115-016-1015-z}
	
	\bibitem[{Tenenbaum(2000)}]{tenenbaum2000}
	Tenenbaum JB (2000) {A Global Geometric Framework for Nonlinear Dimensionality
		Reduction}. Science 290(5500):2319--2323, \doi{10.1126/science.290.5500.2319}
	
	\bibitem[{Torgerson(1952)}]{Torgerson1952}
	Torgerson W (1952) {Multidimensional scaling: I. Theory and method}.
	Psychometrika 17(4):401--419, \doi{10.1007/BF02288916}
	
	\bibitem[{{Van Der Maaten} et~al(2009){Van Der Maaten}, Postma, and {Van Den
			Herik}}]{Laurens2009}
	{Van Der Maaten} LJP, Postma EO, {Van Den Herik} HJ (2009) {Dimensionality
		Reduction: A Comparative Review}. Journal of Machine Learning Research
	10:1--41, \doi{10.1.1.112.5472}
	
	\bibitem[{Van Der~Putten and Van~Someren(2004)}]{coil}
	Van Der~Putten P, Van~Someren M (2004) A bias-variance analysis of a real world
	learning problem: The coil challenge 2000. Machine learning 57(1):177--195
	
	\bibitem[{Venna(2007)}]{Venna2007}
	Venna J (2007) {Dimensionality reduction for visual exploration of similarity
		structures}. Helsinki University of Technology
	
	\bibitem[{Vergara et~al(2012)Vergara, Vembu, Ayhan, Ryan, Homer, and
		Huerta}]{batch}
	Vergara A, Vembu S, Ayhan T, Ryan MA, Homer ML, Huerta R (2012) {Chemical gas
		sensor drift compensation using classifier ensembles}. Sensors and Actuators
	B: Chemical 166:320--329, \doi{10.1016/j.snb.2012.01.074}
	
	\bibitem[{Vincent et~al(2008)Vincent, Larochelle, Bengio, and
		Manzagol}]{vincent2008extracting}
	Vincent P, Larochelle H, Bengio Y, Manzagol PA (2008) Extracting and composing
	robust features with denoising autoencoders. In: Proceedings of the 25th
	International Conference on Machine Learning, ACM, pp 1096--1103,
	\doi{10.1145/1390156.1390294}
	
	\bibitem[{Wang et~al(2014)Wang, Huang, Wang, and Wang}]{Wang2014}
	Wang W, Huang Y, Wang Y, Wang L (2014) {Generalized autoencoder: A neural
		network framework for dimensionality reduction}. In: Proceedings of the IEEE
	Conference on Computer Vision and Pattern Recognition Workshops, pp 496--503,
	\doi{10.1109/CVPRW.2014.79}
	
	\bibitem[{Wang(2011)}]{wang2012}
	Wang X (2011) A fast exact k-nearest neighbors algorithm for high dimensional
	search using k-means clustering and triangle inequality. In: Proceedings of
	the International Joint Conference on Neural Networks, IEEE, pp 1293--1299,
	\doi{10.1016/j.patcog.2010.01.003}
	
	\bibitem[{Weinberger and Saul(2006)}]{weinberger2006}
	Weinberger KQ, Saul LK (2006) An introduction to nonlinear dimensionality
	reduction by maximum variance unfolding. In: Proceedings of the 21st National
	Conference on Artificial Intelligence, vol~2, pp 1683--1686
	
	\bibitem[{Wiley(2016)}]{willey2016}
	Wiley JF (2016) {R Deep Learning Essentials}. Packt Publishing Ltd.
	
	\bibitem[{Xia et~al(2017)Xia, Peng, Zhang, and Bae}]{xia2017demst}
	Xia Y, Peng Y, Zhang X, Bae H (2017) Demst-knn: A novel classification
	framework to solve imbalanced multi-class problem. In: Computer Science
	On-line Conference, Springer, pp 291--301,
	\doi{10.1007/978-3-319-57261-1\_29}
	
	\bibitem[{Yang et~al(2011)Yang, Zhang, Yang, and Zhang}]{yang2011robust}
	Yang M, Zhang L, Yang J, Zhang D (2011) Robust sparse coding for face
	recognition. In: IEEE Conference on Computer Vision and Pattern Recognition,
	IEEE, pp 625--632, \doi{10.1109/CVPR.2011.5995393}
	
	\bibitem[{Ye et~al(2018)Ye, Chen, Hou, Hardy, and Li}]{Ye2018}
	Ye Y, Chen L, Hou S, Hardy W, Li X (2018) Deepam: a heterogeneous deep learning
	framework for intelligent malware detection. Knowledge and Information
	Systems 54(2):265--285, \doi{10.1007/s10115-017-1058-9},
	\urlprefix\url{https://doi.org/10.1007/s10115-017-1058-9}
	
	\bibitem[{Yu et~al(2001)Yu, Ooi, Tan, and Jagadish}]{yu2001}
	Yu C, Ooi BC, Tan KL, Jagadish HV (2001) {Indexing the Distance: An Efficient
		Method to KNN Processing}. In: International Conference on Very Large
	Databases, pp 421--430
	
	\bibitem[{Yu and Yang(2001)}]{yu2001direct}
	Yu H, Yang J (2001) A direct lda algorithm for high-dimensional data-with
	application to face recognition. Pattern Recognition 34(10):2067--2070,
	\doi{10.1016/S0031-3203(00)00162-X}
	
	\bibitem[{Zhang et~al(2014)Zhang, Tian, Mu, and Fan}]{zhang2014supervised}
	Zhang J, Tian G, Mu Y, Fan W (2014) Supervised deep learning with auxiliary
	networks. In: Proceedings of the 20th ACM SIGKDD international conference on
	Knowledge discovery and data mining, ACM, pp 353--361
	
	\bibitem[{Zhou et~al(2016)Zhou, Zhu, He, and Hu}]{Zhou2016}
	Zhou G, Zhu Z, He T, Hu XT (2016) Cross-lingual sentiment classification with
	stacked autoencoders. Knowledge and Information Systems 47(1):27--44,
	\doi{10.1007/s10115-015-0849-0},
	\urlprefix\url{https://doi.org/10.1007/s10115-015-0849-0}
	
\end{thebibliography}

%\clearpage
\newpage

\textbf{About the author} -- FRANCISCO JAVIER PULGAR RUBIO received his B.Sc. degree in Computer Science from the University of Ja\'en in 2013. He is a PhD student in the Computer Science Department at the University of Ja\'en (Spain). His research interests include areas such as classification, imbalance problems, high dimensionality problem,  neural network design, subgroup discovery, big data and parallelization techniques.

\textbf{About the author} -- FRANCISCO CHARTE received his B.Eng. degree in Computer Science from the University of Ja\'en in 2010 and his M.Sc. and Ph.D. in Computer Science from the University of Granada in 2011 and 2015, respectively. He is currently a researcher at the University of Ja\'en (Spain). His main research interests include machine learning with applications to multi-label classification, high dimensionality and imbalance problems, and deep learning related techniques.

\textbf{About the author} -- ANTONIO J. RIVERA received his B.Sc. degree and his Ph.D. in Computer Science from the University of Granada in 1995 and 2003, respectively. He is a lecturer of Computer Architecture and Computer Technology with the Computer Science Department at the University of Ja\'en (Spain). His research interests include areas such as multilabel classification, imbalance problems, evolutionary computation,  neural network design, time series prediction and regression tasks.

\textbf{About the author} -- MAR\'IA J. DEL JESUS received the M.Sc. and Ph.D. degrees in Computer Science from the University of Granada, Granada, Spain, in 1994 and 1999, respectively. She is an Associate Professor with the Department of Computer Science, University of Ja\'en, Spain. Her current research interests include fuzzy rule-based systems, genetic fuzzy systems, subgroup discovery, data preparation, feature selection, evolutionary radial basis neural networks, knowledge extraction based on evolutionary algorithms, and data mining.

\end{document}